%% file: main.tex
\documentclass[journal,article,submit,moreauthors,pdftex,10pt,a4paper]{Definitions/mdpi} 
\usepackage[utf8]{inputenc}
\usepackage{geometry}
\usepackage{mathtools}
\usepackage{color}
\usepackage{booktabs}
\usepackage{subcaption}
\usepackage{adjustbox}
\usepackage[scale=2]{ccicons}
\usepackage{amssymb}
\usepackage{amsmath}
\usepackage{tikz}
\usepackage{float}
\usepackage{xspace}
\usepackage{comment}
\usepackage{multirow}

\usepackage[font={small}]{caption}

\Title{A Comparison Study on Nonlinear Dimension Reduction Methods with Kernel Variations: Visualization, Optimization and Classification}

\Author{Katherine C. Kempfert$^{1,\dagger}$, Yishi Wang $^{2}$*, Cuixian Chen $^{2}$, and Samuel W.K. Wong $^{3}$}

\AuthorNames{Katherine C. Kempfert, Yishi Wang, Cuixian Chen and Samuel W.K. Wong}

\address{%
$^{1}$ \quad University of Florida; kkempfert2@ufl.edu\\
$^{2}$ \quad University of North Carolina Wilmington; \{wangy, chenc\}@uncw.edu\\
$^{3}$ \quad University of Waterloo;\ samuel.wong@uwaterloo.ca}

\corres{Correspondence: wangy@uncw.edu; Tel.: +1-910-962-3292}

\firstnote{Current address: The University of Florida. Gainesville, FL 32611} 

\abstract{
Because of high dimensionality, correlation among covariates, and noise contained in data, dimension reduction (DR) techniques are often employed to the application of machine learning algorithms. Principal Component Analysis (PCA), Linear Discriminant Analysis (LDA), and their kernel variants (KPCA, KLDA) are among the most popular DR methods. Recently, Supervised Kernel Principal Component Analysis (SKPCA) has been shown as another successful alternative. In this paper, brief reviews of these popular techniques are presented first. We then conduct a comparative performance study based on three simulated datasets, after which the performance of the techniques are evaluated through application to a pattern recognition problem in face image analysis. The gender classification problem is considered on MORPH-II and FG-NET, two popular longitudinal face aging databases. Several feature extraction methods are used, including biologically-inspired features (BIF), local binary patterns (LBP), histogram of oriented gradients (HOG), and the Active Appearance Model (AAM). After applications of DR methods, a linear support vector machine (SVM) is deployed with gender classification accuracy rates exceeding 95\% on MORPH-II, competitive with benchmark results. A parallel computational approach is also proposed, attaining faster processing speeds and similar recognition rates on MORPH-II. Our computational approach can be applied to practical gender classification systems and generalized to other face analysis tasks, such as race classification and age prediction. 
}


\begin{document}
\maketitle
\section{Introduction}
Due to advances in data collection and storage capabilities, the demand has been growing substantially for gaining 
insights into high-dimensional, complex-structured, and noisy data. Researchers from diverse areas have applied DR techniques to visualize and analyze such data \cite{fodor2002survey, sorzano2014survey}. DR techniques are also helpful to address the issues of collinearity and $"p\gg n"$ (i.e., number of features exceeding the sample size in a dataset), by projecting the data into a lower dimensional space with less correlation, so that classical statistical methods can be applied \cite{izenman2008modern}. Principal Component Analysis (PCA) \cite{pearson1901liii, hotelling1933analysis} is a well-studied algorithm with the goal of projecting input features onto a lower dimensional subspace while preserving the largest variance possible; lower dimensionality permits easier visualization, for example via heat maps. While PCA is a fully automatic algorithm, DR techniques that account for domain expertise via user input have also been more recently studied \cite{yang2003interactive, johansson2009interactive}. For classification problems, in which the label information as the response variable is available, Linear Discriminant Analysis (LDA) (sometimes referred to as Fisher's Discriminant Analysis (FDA)) can be used for DR by minimizing intra-class variation and maximizing inter-class variation \cite{fisher1936use, rao1948utilization}. Since PCA only utilizes the correlation or covariance matrices, it is considered an unsupervised approach, whereas LDA is considered a supervised approach with labeling information built into its objective function. Despite the dissimilarities, both PCA and LDA search for linear combinations of the features and, therefore, can be applied in linearly separable types of problems \cite{lee2007nonlinear}.
The main challenge is that many problems in practical applications of machine learning are nonlinear \cite{nhan2015beyond,yin2012kernel}. For nonlinear DR, kernel methods are popular choices because of their flexibility \cite{shawe2004kernel, motai2015kernel,xie2012robust}, e.g, Kernel PCA \cite{scholkopf1997kernel}, Kernel LDA for two classes \cite{mika1999fisher}, and more generalized Kernel LDA for multiple classes \cite{baudat2000generalized}. For kernel methods, it is also possible to design specialized kernels based on domain knowledge of a problem \cite{barzilay1999domain, scholkopf1998prior}.


Given the problems in image analysis of high dimensionality and complex correlation structures, DR techniques are often a necessary step \cite{hinton2006reducing}. 
Thus, variants of PCA, LDA, and their kernel extensions have been popular in computer vision with applications of image classification and discrimination \cite{zhao1998discriminant,martinez2001pca,yang2003can}. Studies include Eigenfaces \cite{turk1991eigenfaces}, Fisherfaces \cite{belhumeur1997eigenfaces}, face recognition with KPCA \cite{kim2002face}, face recognition with Kernel Direct LDA \cite{lu2003face}, 2D-PCA \cite{yang2004two}, 2D-LDA \cite{li20052d}, among many others.
When there are sufficient labeled face images, LDA is experimentally reported to outperform PCA for face recognition \cite{belhumeur1997eigenfaces}. In the case of a small training set, the conclusion could be reversed \cite{martinez2001pca}. Studies comparing classification performance of PCA, LDA, and their kernel variations include \cite{karg2009comparison, ye2009comparative}. The connections among KLDA, KPCA, and LDA are further discussed in \cite{yang2004essence}.
By incorporating labeling information into the construction of the objective function, Supervised Kernel PCA (SKPCA) \cite{barshan2011supervised} has been proposed for visualization, regression, and classification. A modified version of SKPCA for classification problems can be found in \cite{wang2015modified}. These studies suggest that SKPCA works well in practice among different DR algorithms \cite{fewzee2012dimensionality,samadani2013discriminative,wu2013biomedical}. Moreover, it has been found in \cite{ashtiani2015dimension} that with bounded kernels, projections from SKPCA are uniformly converging, regardless of the input features' dimension.



\section{Associated Work}
In recent years, facial demographic analysis has become popular in computer vision, because of its broad applications in human-computer interaction (HCI), security, surveillance, and marketing, which can benefit from the automatic estimation of characteristics like age, gender, and race. Recent surveys on demographic estimation from biometrics are presented in \cite{fu2010age, sun2018demographic}.
Specifically, a major task is gender classification, aiming to automatically determine if a person is male or female. Beyond computer vision, the topic has been studied extensively by anthropologists, sociologists, and psychologists. Gender can easily be identified by humans, achieving 96\% accuracy in an experiment classifying photographs of adult faces \cite{burton1993s}. Automating gender classification has been a priority in real-world  applications. 
A number of biometrics have been used to identify gender, including face, voice, gait, handwriting, and even the iris \cite{sun2018demographic}. However, gender classification from faces is the most common, probably because photography of faces is non-intrusive and ubiquitous. Ng et al. provide a survey of gender classification via face and gait \cite{ng2012vision}.

Gender classification with faces launched in 1990, when neural networks were applied directly to pixels from face photographs \cite{golomb1990sexnet, cottrell1991empath}. Many other early studies utilized the \textit{geometric-based} approach to represent human faces, relying on measurements of facial landmarks \cite{poggio1992hyberbf, wiskott1995face}. Though intuitive, such approaches are sensitive to the placement of landmarks, which can only accommodate frontal representations of the face, and may omit some important information from the face (such as texture of the skin). In recent years, the \textit{appearance-based} methods have been more commonly adopted, which rely on a transformation of an image's pixels \cite{guo2010human,guo2011simultaneous,shan2012learning}. Such methods capture both the geometric relationships of the face and texture information. However, a drawback is their sensitivity to illumination and viewpoint variations. Other issues are associated with the high dimensionality of the transformed pixels, which will be discussed further in the next paragraph. Some most recent gender classification studies involve convolutional neural networks (CNN) \cite{yang2013automatic,yi2014age,antipov2016minimalistic,antipov2017effective}. Though CNNs have reached state-of-the-art accuracy rates, they are known to be less interpretable than some other methods.


Pixels often contain high redundancy and noise, which cannot be removed completely by pre-processing steps. Hence, the vectors resulting from \textit{appearance-based} feature extraction methods genetically inherit redundancy and noise. Popular image feature extraction methods include local texture techniques such as local binary patterns (LBP) \cite{yang2007demographic,lian2006multi,makinen2008experimental,alexandre2010gender}, Gabor filters \cite{xia2008multi}, biologically-inspired features (BIF) \cite{guo2009gender, han2015demographic}, and histogram of oriented gradients (HOG) \cite{guo2009gender}. Such methods 
could lead to a high dimension of extracted features, thwarting practical applications by increasing runtime and memory consumption. 
%
When "$p\gg n$", for which the dimension of the feature space exceeds the sample size of the dataset, a fundamental assumption of many standard statistical procedures is violated. Additionally, collinearity of features can cause numerical problems, while noisy features can obscure true relationships with the response variable and hinder predictive performance. These significant issues motivate the use of DR techniques. The fundamental goal of DR is to extract and retain information in a lower dimensional space. Many of these methods fall under manifold learning, identifying a low-dimensional manifold embedded in a high-dimensional ambient space \cite{ma2011manifold}.

Even though PCA and LDA have been widely considered as popular and effective approaches for DR in machine learning, their kernel versions are much less investigated. To our best knowledge, KPCA, KLDA, and SKPCA have never before been directly compared on visualization and classification performance through simulations and practical applications to face image analysis problems. 

Our main contributions in this study can be summarized as follows. (1) The nonlinear manifold learning projections for KPCA, KLDA, and SKPCA are directly compared with visualization through simulated datasets. (2) Motivated by the nonlinear nature of soft-biometric analysis problems, we utilize KPCA, KLDA, and SKPCA for dimension reduction on four types of appearance-based extracted features (BIF, HOG, LBP, and AAM) for the gender classification task. Moreover, the classification performance is compared systematically on parameter optimization. (3) For applications to practical large-scale systems, we propose an additional parallel computational framework that can decrease runtime while maintaining similar classification rates. 

The remainder of the paper is structured as follows. In Section 3, we review the theory of KPCA, SKPCA, and KLDA. In Section 4, we conduct simulation studies to visualize projections. We propose our machine learning methods for gender classification on Morph-II in Section 5. The comparative performance of KPCA, SKPCA, and KLDA on Morph-II is presented and discussed in Section 6. The performance of these DR methods is further compared in Section 7 through application to the FG-NET dataset. The computational framework for large-scale practical systems is proposed in Section 8 and investigated on Morph-II. Finally, we conclude and offer future directions of research in Section 9.

\section{Kernel-Based Dimension Reduction Methods}

The nonlinearity in a classification problem can often be addressed by kernel-based DR methods, with the appropriate choice of kernels. The driving reasons are the nonlinearity of chosen kernels, flexibility of tuning parameter selection, and most importantly, the kernel tricks. Mercer's theorem guarantees that a symmetric positive-definite function can be written as the sum of a convergent sequence of product functions, which potentially project the data into infinite-dimensional space \cite{scholkopf2001generalized}. Thus, it is feasible to separate the data in the new space. On the other hand, Representer Theorem shows that the solution for certain kernel methods lies in the finite-dimensional span of the training data \cite{wahba1990spline, scholkopf2001generalized}. This is very helpful, since we do not need to compute the coordinates of the projected data in the infinite-dimensional space, but only the inner products between all pairs of data in the feature space.





\subsection{Notations}
With the goal of emphasizing the connections between KPCA, SKPCA, and KLDA, we define the following notations for classification problems.

Let $\mathcal{X}$ be the feature space, a non-empty subset in $\mathbb{R}^p$ with $p$ as the number of covariates for each subject. Let $\mathcal{Y}$ be the space for the response variable, a subset in $\mathbb{R}$. Let $\{(x_1,y_1),\cdots, (x_n,y_n)\}\subset\mathcal{X}\times\mathcal{Y} $ be a series of $n$ independent observations following a joint probability measure $P_{\mathcal{X}, \mathcal{Y}}$. Let $Y =[y_{1}, y_{2},  \cdots,  y_{n}]^T$ denote the outcomes of the response variable. Let $X$ be an $n\times p$ feature matrix, with $x_{i}^T$ as the $i$-th row for $i=1,\cdots, n$ , and $x^{(l)}\in\mathbb{R}^n$ for $l=1,\cdots, p$ as its $l$-th column. Thus, the $X$ matrix can be written as:
\begin{equation*}
{{X}}=
\begin{bmatrix}
x_{1}, x_{2},  \cdots,  x_{n}
\end{bmatrix}^{T}=
\begin{bmatrix}
x^{(1)}, x^{(2)},  \cdots,  x^{(p)}
\end{bmatrix}.
\end{equation*}
Without loss of generality, we may assume that each column of the $X$ matrix is normalized, such that the mean of $x^{(l)}$ is 0 and standard deviation is 1, for $l=1,\cdots,p$.

Let $\Sigma$ be the sample covariance matrix of $X$. 
We then have
\begin{equation}\label{def:sig}
\hspace{1cm}
\underset{p\times p}{{\Sigma}} = \frac{1}{n-1}X^TX=
\frac{1}{n-1}\sum_{i=1}^{n}{x_{i}{x_{i}}^{T}}.
\end{equation}

%
Let $\mathcal{F}$ be a reproducing kernel Hilbert space on $\mathcal{X}$ from a kernel function $k(\cdot,\cdot)$, which is a Mercer  kernel (symmetric and positive-definite), and $\mathcal{G}$ be a reproducing kernel Hilbert space on $\mathcal{Y}$ from a kernel function $l(\cdot,\cdot)$.

For the kernel $k: \mathcal{X}\times\mathcal{X} \rightarrow \mathbb{R}$, its associated space $\mathcal{F}$ may be infinite-dimensional, but with some additional conditions, the minimizer of a regularized risk function lies in the finite span of the training observations \cite{scholkopf2001generalized}. Additionally, it has been shown \cite{scholkopf2001generalized} that there exists a function 
\begin{equation}
\phi: \mathcal{X} \rightarrow \mathcal{F}
\end{equation} such that for all $x, x' \in \mathcal{X} $,
\begin{equation}
\label{eq:kernel}
k(x,x')=<\phi(x),\phi(x')>,
\end{equation}
where $<\cdot>$ is the dot product. 
Let 
$K$ be a matrix such that its $ij$-th element is $k(x_i,x_j)$. We then have
\begin{align}\label{def:K}
K=\{k(x_i,x_j)\}_{ij}=\{<\phi(x_i),\phi(x_j)>\}_{ij}=\Phi(X)\Phi(X)^T,
\end{align}
where $\Phi(X)=[\phi(x_1), \phi(x_2), \cdots, \phi(x_n)]^T$. Here, the kernel matrix $K$ is the Gram matrix of the $\phi(x_i)$'s.

\subsection{Principal Component Analysis and Kernel Principal Component Analysis}

In standard PCA, we seek an orthogonal transformation matrix $A$ satisfying
\begin{equation}
\underset{n\times d}{{T}}=\underset{n\times p}{{X}}\hspace{.2cm}\underset{p\times d}{{A}},
\end{equation}
where $T=[t_1, t_2,\cdots, t_d]$ for some $d \leq p$, such that each column vector $t_{i}$ successively inherits maximal proportion of variance from the column vectors $x^{(l)}$'s, while ensuring the projection directions are perpendicular.
The solutions can be expressed as the eigenvalue problem
\begin{equation}
\Sigma a_i=\lambda_i a_i,
\end{equation}
where $a_i$ is the $i$-th column of $A$, for $i=1,\hdots,d$.

Following the work of \cite{scholkopf1998nonlinear}, PCA can be extended to KPCA by first choosing a Mercer kernel $k$, with which $x_i$ is transformed to $\phi(x_i)$. This maps the features in $X$ to $\Phi(X)$. Assume that $\sum_{i=1}^n\phi(x_i)$ is a vector with 0 in each entry. With the Gram matrix $K=\Phi(X)\Phi(X)^T$ as defined in  (\ref{def:K}) and through the kernel trick from (\ref{eq:kernel}), 
we have the eigenvalue problem 

\begin{equation}
\label{eigprob}
{K}a_{i}^{*} = \lambda_i ^{*} a_{i}^{*},
\end{equation}
where $d$ is the desired dimension and $a_{i}^{*},\cdots,a_{d}^{*}$ are the eigenvectors of $K$, with associated eigenvalues $\lambda _{1}^{*} \geq  \lambda_{2}^{*} \geq \cdots\geq\lambda_{d}^{*}$.
Hence, the advantage of the kernel-based approach is to calculate the Gram matrix $K$ without an explicit expression for $\phi$. 
Without the centralization assumption on $\phi$, the $K$ matrix in (\ref{eigprob}) can be replaced by 
\begin{equation}\label{kalter}
{K}^{*} = H_nKH_n, 
\end{equation}
where $d$ is the desired dimension, $H_n=I_n-\frac{1}{n}1_n$, $I_{n}$ is an identity matrix with dimension $n\times n$, and $1_{n}$ is a matrix of 1's with dimension $n\times n$. 

We note that $H_n$ is idempotent, since it is a square matrix satisfying $H_n=H_nH_n$. For any square matrix $S$ with dimension $n \times n$, the average of each column of the matrix $H_nS$ is 0, as is the average of each row of the matrix $SH_n$. Thus, the $K^*$ matrix is the "centralized" version of the original $K$ matrix. 

\subsection{Supervised Kernel Principal Component Analysis}
PCA and KPCA are unsupervised methods, since they do not consider the response variable, only considering directions of maximum variability in the covariates. If the goal is classification, this may not be ideal, since the principal components may be unrelated to the class difference.
SKPCA is a supervised generalization of KPCA, which aims to find the principal components with maximal dependence on the response variable. Drawing from \cite{barshan2011supervised} and \cite{wang2015modified}, we formulate SKPCA as follows.

In SKPCA, class information is incorporated by maximizing the Hilbert Schmidt independence criterion (HSIC) \cite{gretton2005measuring}. With the aforementioned reproducing kernel Hilbert spaces $\mathcal{F}$ on $\mathcal{X}$ and $\mathcal{G}$ on $\mathcal{Y}$ and related kernel functions $k(\cdot,\cdot)$ and  $l(\cdot,\cdot)$ respectively, the HSIC can be expressed as 
\begin{equation}
\label{eq:HSIC}
\begin{aligned}
HSIC(P_{\mathcal{X},\mathcal{Y}},\mathcal{F},\mathcal{G})=E_{x,x',y,y'}[k(x,x')l(y,y')]&+E_{x,x'}[k(x,x')]E_{y,y'}[l(y,y')]\\
&-2E_{x,y}\big(E_{x'}[k(x,x')]E_{y'}[{l(y,y')}]\big),
\end{aligned}
\end{equation}
where $E_{x,x',y,y'}$ represents the expectation on independent pairs of $(x, y)$ and $(x', y')$ (with respect to $P_{\mathcal{X},\mathcal{Y}}$) and $E_{x,x'}$ and alike are the expectations based on various marginal distributions from $P_{\mathcal{X},\mathcal{Y}}$.

With the results from \cite{gretton2005measuring}, an empirical estimator of (\ref{eq:HSIC}) is
\begin{equation}
\label{eq:HSIC_est}
HSIC(X,Y,\mathcal{F},\mathcal{G})=\frac{1}{(n-1)^{2}}tr(KH_nLH_n),
\end{equation}
where 
$K$ and $H_n$ are defined as before for KPCA and $L=\{1(y_{i}=y_{j})\}_{ij}$ is a link matrix with dimension $n \times n$, where $1(\cdot)$ is an indicator function with value 1 if the event is true and 0 otherwise. 

Similarly as for KPCA, $K$ and $L$ can be adjusted to satisfy the centralization assumption. As discussed previously, $H_n$ is an idempotent matrix. Therefore, following from (\ref{eq:HSIC_est}),
\begin{align}\label{eq:HSIC_est2}
HSIC^*(X,Y,\mathcal{F},\mathcal{G})=&\frac{1}{(n-1)^{2}}tr(KH_nH_nLH_nH_n)\nonumber\\
=&\frac{1}{(n-1)^{2}}tr(H_nKH_nH_nLH_n)\nonumber\\
=&\frac{1}{(n-1)^{2}}tr(K^*L^*),
\end{align}
where $K^*$ and $L^*$ are the "centralized" versions of the $K$ and $L$ matrices respectively.

On another note, in the binary gender classification problem, $\textrm{rank}(L)=2$ and $\textrm{rank}(KH_nLKH_n)\leq 2$ \cite{wang2015modified}.
Therefore, we modify the link matrix according to \cite{wang2015modified} by
\begin{equation}
\label{eq:improved_link}
L=\{1(y_{i}=y_{j}) \times k(x_{i},x_{j})\}_{ij}.
\end{equation}

It can be shown that maximization of (\ref{eq:HSIC_est}) is equivalent to solving the generalized eigenvalue problem
\begin{equation}
\label{eq:SKPCA_eigen}
Av_i=\lambda_i Kv_i,
\end{equation}
where $A=KH_nLH_nK$ and each $v_i$ is an eigenvector with related eigenvalue $\lambda_i$ for $i = 1, \cdots, d$, where $d$ is the desired dimension \cite{wang2015modified}.
Therefore, the main advantage of the link matrix in (\ref{eq:improved_link}) becomes apparent: the rank of $KH_nLKH_n$ may increase, permitting more eigenvalues to be computed. 

\subsection{Linear Discriminant Analysis and Kernel Linear Discriminant Analysis}
Given a dataset with finite classes, LDA aims to find the best set of features to discriminate among the classes. 
We first review standard LDA, then generalize to KLDA. We note that sometimes parametric assumptions for LDA are made, such as that observations from each class are normally distributed with common covariance. Here, we make no such assumptions. 
%
Suppose that each observation $x_{i}$ for $i=1,\cdots,n$ belongs to exactly one of $C$ classes. Define the following feature vectors: $\bar{x}=\frac{1}{n}\sum_{i=1}^nx_i$ as the overall mean and $\bar{x_{c}}=\frac{1}{n_c}\sum_{i=1}^nx_i1(x_i\in \text{class c})$ as the mean of the $c$-th class with $n_{c}$ the size of the $c$-th class in the sample, for $c=1,\cdots,C$. 

In standard LDA, we seek to maximize the objective function
\begin{equation}
\label{J_kfda}
J(v)=\frac{v^{T}S_{B}v}{v^{T}S_{W}v},
\end{equation}
where $v$ is a $p$ x $1$ vector, $S_{B}$ is the between-class scatter matrix, and $S_{W}$ is the within-class scatter matrix defined by
\begin{equation}
\begin{aligned}
\underset{p\times p}{S_{B}}&=\sum_{c}n_{c}{(\bar{x_{c}}-\bar{x}){(\bar{x_{c}}-\bar{x})}^{T}} \textrm{ and}\\
\underset{p\times p}{S_{W}}&=\sum_{c}\sum_{i\in c}{(x_{i}-\bar{x_{c}}){(x_{i}-\bar{x_{c}})^{T}}}.
\end{aligned}
\end{equation}
Hence, maximizing $J(v)$ involves finding some rotation of the scatter matrices such that the "distance" between the groups is maximized relative to the variations within each group.

Maximization of $J(v)$ in (\ref{J_kfda}) is equivalent to solving the generalized eigenvalue problem
\begin{equation}
\label{ga:lda}
S_{B}v_i=\lambda_i {S_{W}}v_i,
\end{equation}
where each $v_i$ is an eigenvector with corresponding eigenvalue $\lambda_i$, for $i = 1, \cdots, d$, where $d$ is the desired dimension.

LDA is generalized to KLDA using the kernel representation from (\ref{eq:kernel}).
Analogously to LDA above, we seek a solution $v^*$ that will result in the maximization of the objective function
\begin{equation}\label{v:klda}
J(v)=\frac{v^TS_B^*v}{v^TS_w^*v},
\end{equation}
where now $v\in \mathcal{F}$ and $S_B^*$ and $S_W^*$ are the between-class and within-class scatter matrices in $\mathcal{F}$ defined by
\begin{equation}
\begin{aligned}
m^{\phi}&=\frac{1}{n}\sum_{i=1}^n\phi(x_i),\\
m_c^{\phi}&=\frac{1}{n_c}\sum_{i=1}^n\phi(x_i)1(x_i\in \text{class c}),\\
{S_{B}^*}&=\sum_{c}n_{c}{(m_c^{\phi}-m^{\phi}){(m_c^{\phi}-m^{\phi})}^{T}}, \textrm{ and} \\
{S_{W}^*}&=\sum_{c}\sum_{i\in c}{(\phi(x_{i})-m_c^{\phi}){(\phi(x_{i})-m_c^{\phi})^{T}}}.
\end{aligned}
\end{equation}

The above expressions involve knowledge of $\phi$, which is often not available. It can be shown that equation (\ref{v:klda}) is equivalent to
\begin{equation}\label{u:klda}
\begin{aligned}
J(u)&=\frac{u^TMu}{u^TNu},
\end{aligned}
\end{equation}
where
\begin{equation}\label{u:klda2}
\begin{aligned}
M_c&=(M_{cj})_j=\left(\frac{1}{n_c}\sum_{h=1}^nk(x_j,x_h)1(x_h\in \text{class c})\right)_j,\\
\bar{M}&=(\bar{M}_{j})_j=\left(\frac{1}{n}\sum_{h=1}^nk(x_j,x_h)\right)_j,\\
M&=\sum_{c}n_{c}{(M_c-\bar{M}){(M_c-\bar{M})}^{T}},\\
K_c&=K\times outer(X, X_c), \\
N&=\sum_{c}K_c H_{n_c}K_c^T,
\end{aligned}
\end{equation}
$X_c$ is a matrix of dimension $n_c\times p$ with rows being features from the $c$-th class, and $outer(X, X_c)$ is an $n\times n_c$ matrix with its $ij$-th element as $1(x_i \text { is the $j$-th observation in class c} )$. A full discussion of KLDA can be found in \cite{mika1999fisher}.
 
Maximization of $J(u)$ in equation (\ref{u:klda}) is equivalent to solving the generalized eigenvalue problem
\begin{equation}
\label{ga:klda}
Mu_i=\lambda_i Nu_i,
\end{equation}
where each $u_i$ is an eigenvector with associated eigenvalue $\lambda_i$, for $i = 1, \cdots, d$ with $d$ as the desired dimension.

Comparing the generalized eigenvalue problems in (\ref{ga:lda}) and (\ref{ga:klda}), the structures of matrices $S_B$ and $M$ are similar, since both "measure" the variation between different classes.

Let $W_c=[w_{c,1},\cdots,w_{c,n_c}]=K_c H_{n_c}$, a matrix of dimension $n\times n_c$. Due to the centralization function of $H_{n_c}$, $W_c$ has row-sum equal to zero for every row. Besides, $K_c H_{n_c}(K_c H_{n_c})^T=W_cW_c^T=\sum_{i=1}^{n_c}w_{c,i}w_{c,i}^T$. For the matrix $N$, due to the idempotent property of $H_{n_c}$, \begin{align}
N=\sum_{c}K_c H_{n_c}H_{n_c}K_c^T=\sum_{c}\sum_{i=1}^{n_c}w_{c,i}w_{c,i}^T.
\end{align}
Thus, the matrix $N$ has an identical structure to the $S_W$ and $S_W^*$ matrices, which "measure" the overall variation within groups.

\section{Visualization on Simulation Studies}
To visualize and improve understanding of the manifold learning methods KPCA, SKPCA, and KLDA, we apply them in three simulation studies. For comparison, the linear techniques PCA and LDA are also considered. Each dataset contains nonlinear patterns, and the goal is to transform the data to be linearly separable. For this reason, the radial basis function (RBF)
\begin{equation}
k(x_{i},x_{j})=e^{-\delta {||x_{i}-x_{j}||}_2^{2}}
\end{equation}
is chosen as a kernel for each pair of observed vectors $x_i,x_j$. For each DR method, a range of values for the tuning parameter $\delta$ are tested and selected to visually separate the classes. A full discussion of the RBF kernel, among others, can be found in \cite{steinwart2008support}.
Figures \ref{winechoc}, \ref{tr}, and \ref{swiss} compare the original data to 2-dimensional projections from each DR method. In each plot, color corresponds to the true class to which an observation belongs.

\renewcommand{\tablename}{Figure}
\captionsetup[table]{position=bottom}

\begin{table}[H]
\centering
	\begin{tabular}{ccc}
\toprule
\includegraphics[height=1.35in,keepaspectratio]{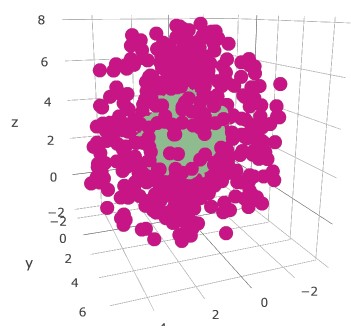} &\includegraphics[height=1.35in,keepaspectratio]{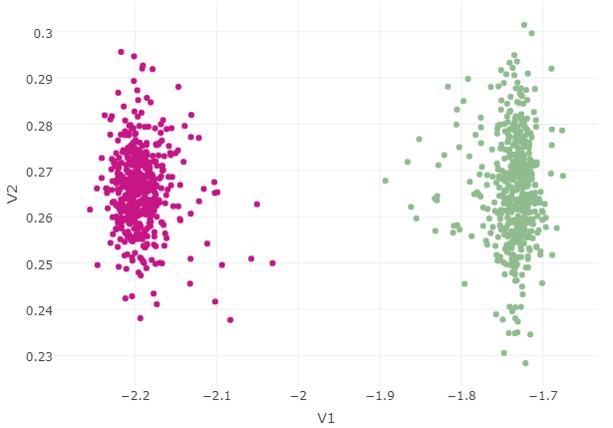} 
&\includegraphics[height=1.35in,keepaspectratio]{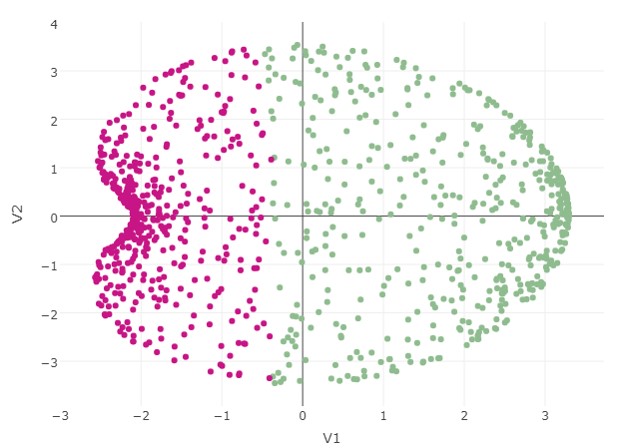}\\
(a) Original Data &(b) KLDA Projections in 2D &(c) KPCA Projections in 2D\\
\addlinespace[4ex]
\includegraphics[height=1.35in,keepaspectratio]{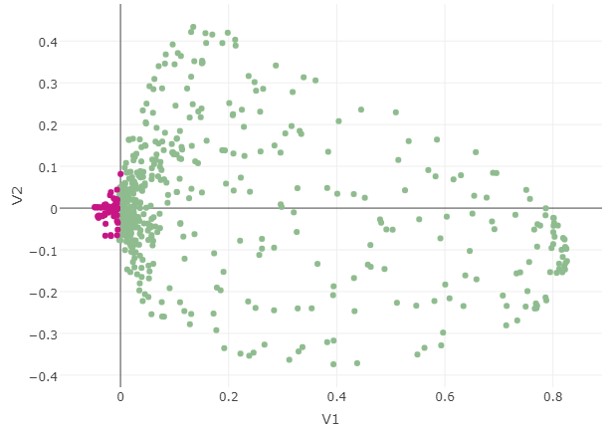}
& \includegraphics[height=1.35in,keepaspectratio]{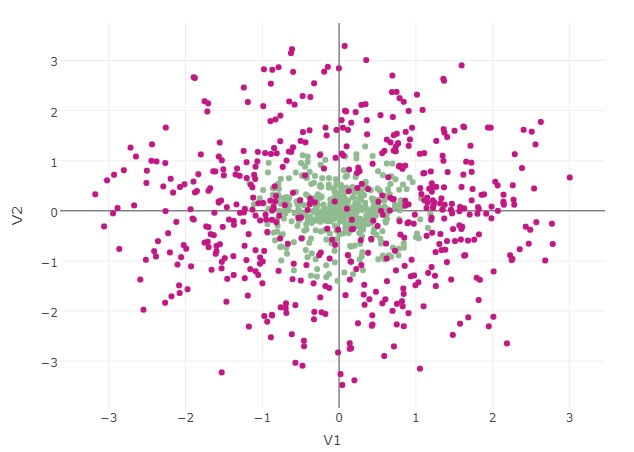}
&\includegraphics[height=1.35in,keepaspectratio]{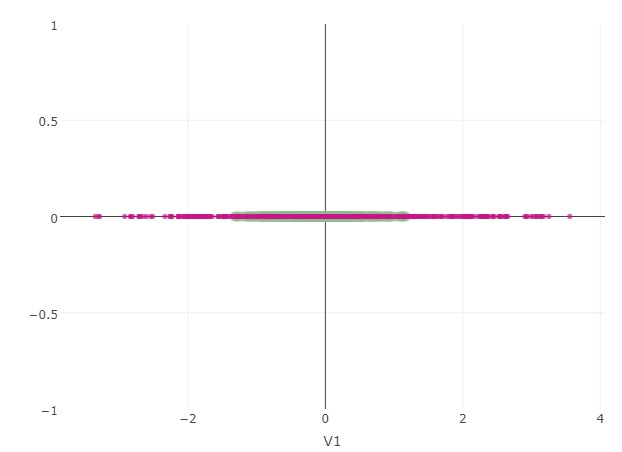}\\
(d) SKPCA Projections in 2D & (e) PCA Projections in 2D & (f) LDA Projections in 2D\\
\bottomrule
\end{tabular}
\captionsetup{justification=justified,singlelinecheck=false}
\caption{Wine Chocolate Simulation Study.}
\label{winechoc}
\end{table}
In the first simulation study, the original data are plotted in 3D in Figure \ref{winechoc}(a); the green sphere is embedded within the magenta group, necessitating nonlinear manifold learning. The KLDA projections in (b) are linearly separable with very good variation between the classes and a fair amount of variation within the classes. KPCA and SKPCA projections in (c) and (d) are at least approximately linearly separable, as it is not clear whether there is a linear boundary that perfectly separates the two classes. In (e), PCA fails to linearly separate the groups, rotating the wine chocolate in 2D. The maximum dimension LDA can retain is $p-1$; with 2 classes, the projections must be plotted on a 1D number line, given in (f). Points from the two classes overlap considerably in plots (e) and (f).

\begin{table}[H]
\centering
	\begin{tabular}{ccc}
\toprule
\includegraphics[height=1.35in,keepaspectratio]{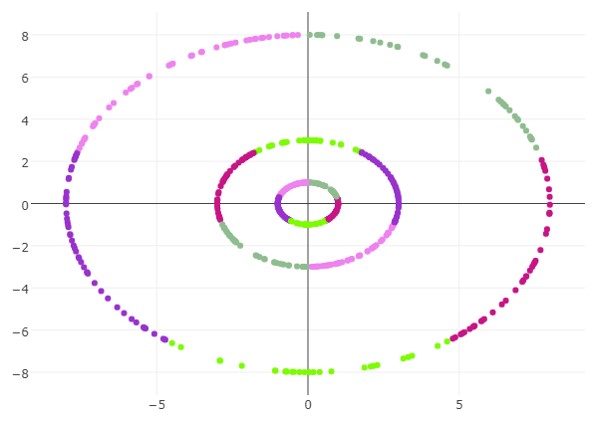} &\includegraphics[height=1.35in,keepaspectratio]{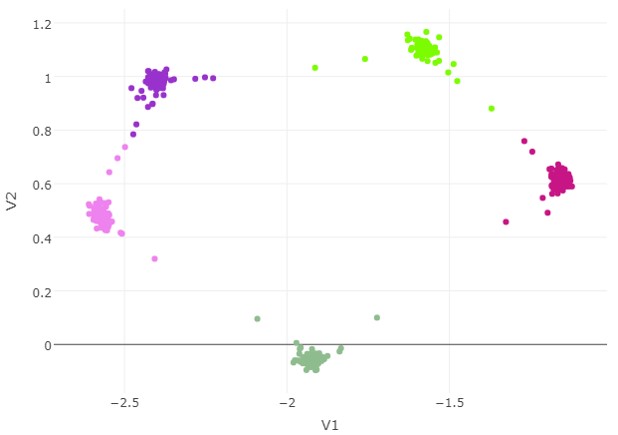} 
&\includegraphics[height=1.35in,keepaspectratio]{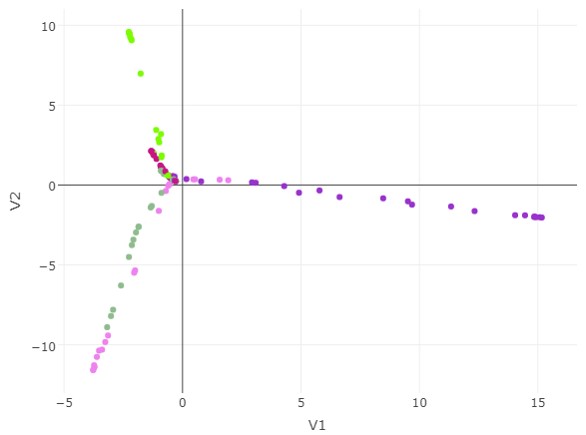}\\
(a) Original Data &(b) KLDA Projections in 2D &(c) KPCA Projections in 2D\\
\addlinespace[4ex]
\includegraphics[height=1.35in,keepaspectratio]{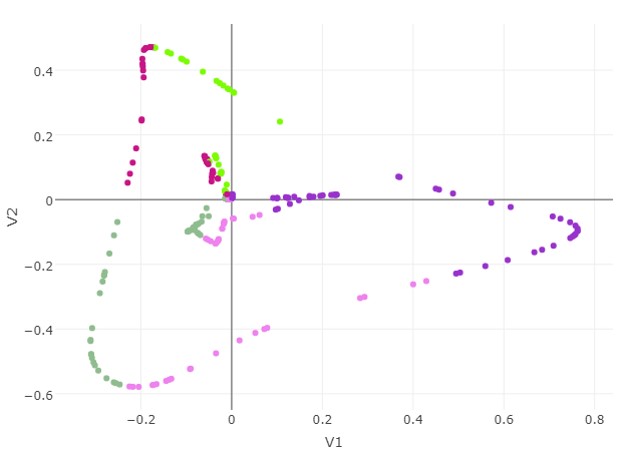}
& \includegraphics[height=1.35in,keepaspectratio]{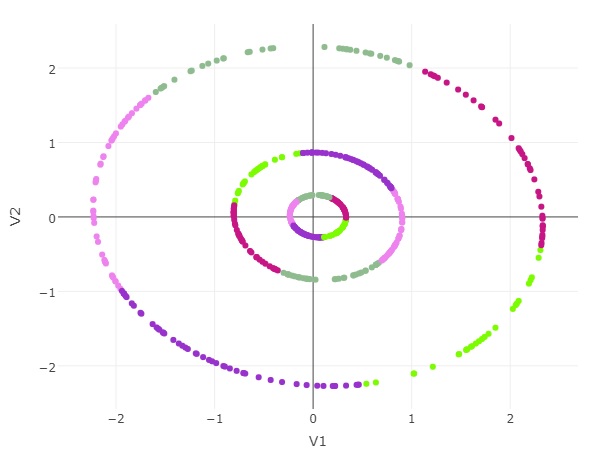}
&\includegraphics[height=1.35in,keepaspectratio]{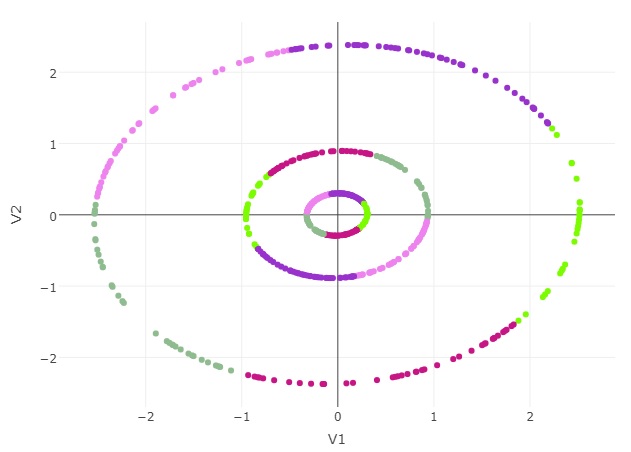}\\
(d) SKPCA Projections in 2D & (e) PCA Projections in 2D & (f) LDA Projections in 2D\\
\bottomrule
\end{tabular}
\captionsetup{justification=justified,singlelinecheck=false}
\caption{Apple Tart Simulation Study.}
\label{tr}
\end{table}
In the second simulation study, the original data in Figure \ref{tr}(a) follow a nonlinear pattern. In (b), KLDA produces groups which are linearly separable. The KPCA projections are approximately linearly separable in (c); however, there is some overlap between groups, especially the green and pink groups in the third quadrant. In (d), SKPCA produces almost linearly separable groups. In plots (e) and (f), PCA and LDA simply rotate the original data in 2D space, as expected.

\begin{table}[H]
\centering
\begin{tabular}{ccc}
\toprule
\includegraphics[height=1.35in,keepaspectratio]{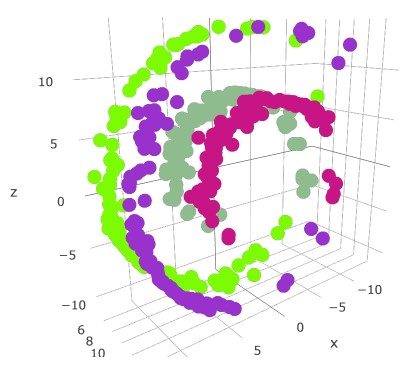} &\includegraphics[height=1.35in,keepaspectratio]{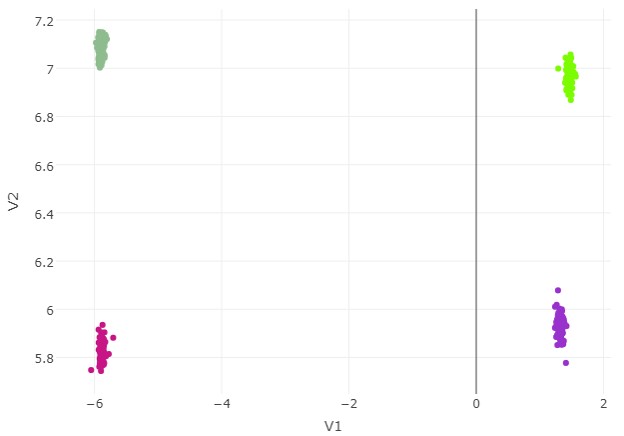} 
&\includegraphics[height=1.35in,keepaspectratio]{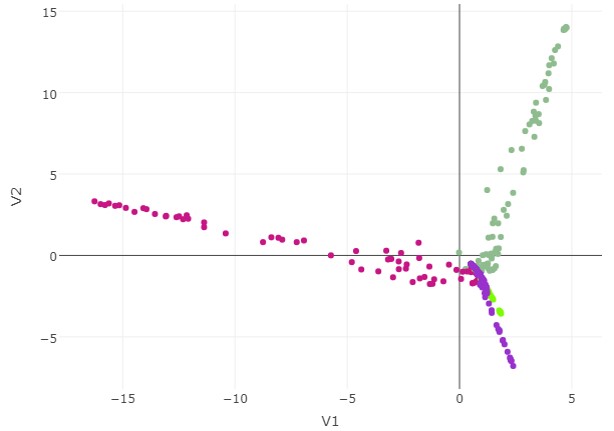}\\
(a) Original Data &(b) KLDA Projections in 2D &(c) KPCA Projections in 2D\\
\addlinespace[4ex]
\includegraphics[height=1.35in,keepaspectratio]{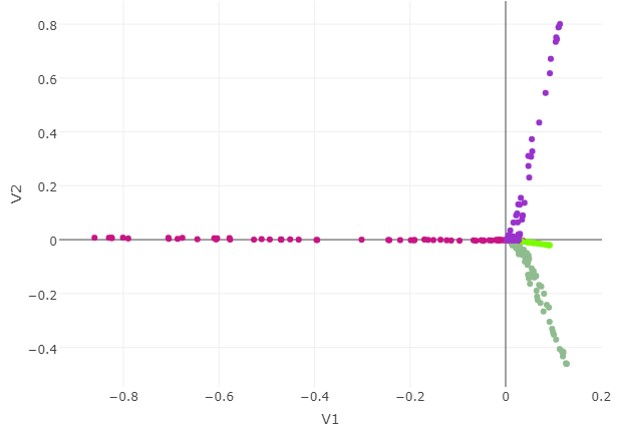}
& \includegraphics[height=1.35in,keepaspectratio]{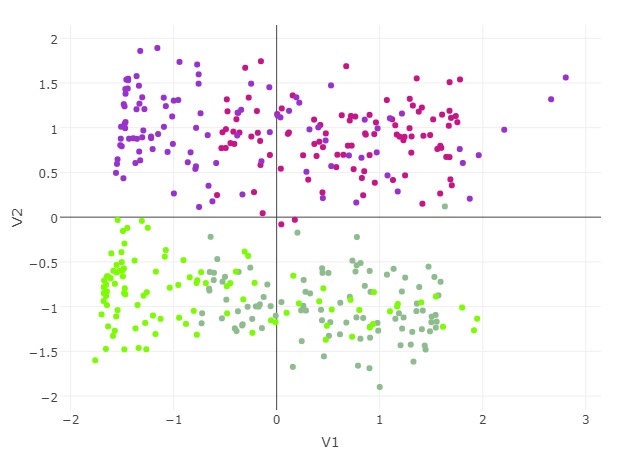}
&\includegraphics[height=1.35in,keepaspectratio]{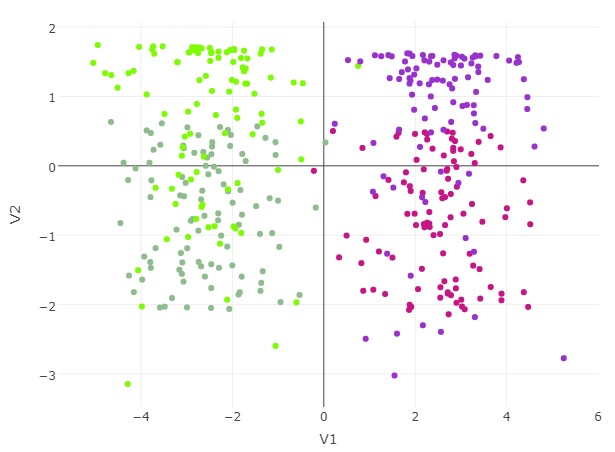}\\
(d) SKPCA Projections in 2D & (e) PCA Projections in 2D & (f) LDA Projections in 2D\\
\bottomrule
\end{tabular}
\captionsetup{justification=justified,singlelinecheck=false}
\caption{Swiss Roll Simulation Study.}
\label{swiss}
\end{table}
For the third simulation study, the original data in Figure \ref{swiss}(a) are in 3D and follow a swirling, nonlinear pattern. In (b), KLDA yields favorable results; the groups are well-separated linearly. KPCA and SKPCA in (c) and (d) also produce good results, although in (c) more separation between the purple and bright green groups would be ideal. In (e) and (f), respectively, PCA and LDA merely rotate the original data projected in 2D space; there is no linear separation between the magenta and purple groups, nor between the two green groups.

\renewcommand{\tablename}{Table}
\captionsetup[table]{position=top}

For all three simulation studies, KLDA, KPCA, and SKPCA are effective to transform the data into linearly separable groups. In all cases, the projected data become approximately linearly separable after applying KLDA, KPCA, or SKPCA. In general, KLDA and SKPCA perform the best here. Their success over KPCA is expected, since KLDA and SKPCA are supervised techniques. On the other hand, results indicate that KPCA and SKPCA are more sensitive than KLDA to different choices of tuning parameters. Hence, SKPCA and KPCA may perform better for alternative choices of parameters. In all our studies, the nonlinear techniques outperform linear PCA and LDA.
These preliminary studies suggest the radial kernel is appropriate for our face analysis experiments.

\section{Kernel-based Dimension Reduction Optimization and Classification on Morph-II}
\noindent Motivated by the nonlinear nature of facial demographic analysis, we propose and implement a novel machine learning process for the Morph-II dataset. We consider the kernel-based DR methods KPCA, SKPCA, and KLDA on three types of appearance-based extracted features (LBP, BIF, and HOG) for the gender classification task. We illustrate parameter optimization and compare the performance of these methods on Morph-II. 

\subsection{Longitudinal Face Database}
MORPH \cite{ricanek2006morph} is one of the most popular face databases available to the public, especially for age estimation, race classification, and gender classification. Multiple versions of MORPH have been released, and the version adopted in this work is the 2008 MORPH-II non-commercial release (referred to as Morph-II in this paper). Morph-II includes over 55,000 mugshots with longitudinal spans and metadata such as date of birth, race, gender, and age.

In addition to its size, Morph-II presents challenges because of disproportionate race and gender ratios. About 84.6\% of images are of males, while only about 15.4\% of images are of females. Imbalanced classes are known to negatively affect certain classification algorithms \cite{japkowicz2002class}. Moreover, Morph-II is skewed in terms of race, with approximately 77.2\% of images picturing black subjects.
Guo et al. found that age, gender, and race interact for demographic analysis tasks including gender classification, race classification, and age prediction \cite{guo2009gender, guo2010study, guo2010human}, so both race and gender imbalance in Morph-II can hamper gender classification.

\subsection{Subsetting Scheme}
To overcome the uneven race and gender distributions in Morph-II, Guo et al.~proposed a subsetting scheme \cite{guo2010human}. Since then, many studies on Morph-II have adopted such an evaluation protocol. 
Based on discussions in Guo et al. \cite{guo2010human}, a new automatic subsetting scheme is proposed in \cite{Yip2018preliminary}, aiming to automatically ensure independent training and testing sets. 
Additionally, inconsistencies in age, gender, and race in Morph-II have been identified and corrected in \cite{Yip2018preliminary}. After following the steps to clean MORPH-II outlined in \cite{Yip2018preliminary}, we apply the automatic subsetting scheme, summarized in Figure \ref{fig:Morph_flow} and described below. 

Let $W$ be the Whole Morph-II dataset, $S$ the selected training/testing set, and $R$ the remaining set. We further divide $S$ into even subsets $S_{1}$ and $S_{2}$. Separately within each subset $S_1$ and $S_2$, we fix the ratios of white (W) to black (B) images as 1:1 and male (M) to female (F) images as 3:1. Further, $S_1$ and $S_2$ have been selected such that the age distributions within each set are similar (details shown in \cite{Yip2018preliminary}). The gender and race summaries for the subsetting scheme are shown in Table \ref{table:images}. Most importantly, the sets $R,S_{1},$ and $S_{2}$ are independent; no sets share images from the same subject. 
We use $S$ as an alternating training and testing set. First, we train on $S_{1}$ and test on $S_{2} \cup R$, then we train on $S_{2}$ and test on $S_{1} \cup R$. The final classification accuracy is obtained by averaging the classification accuracies from the alternations.

\setcounter{figure}{3}  

\begin{figure}
\includegraphics[width=.6\textwidth]{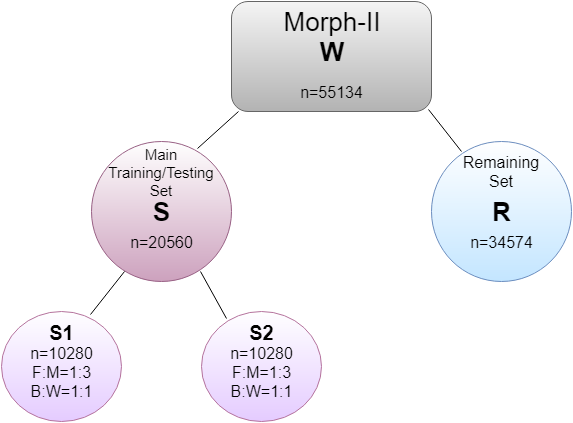}
\centering
\caption{Flowchart representing our subsetting scheme \cite{Yip2018preliminary} for MORPH-II, which improves the one from \cite{guo2010human}.}
\label{fig:Morph_flow}
\end{figure}

\setcounter{table}{1} 

\begin{table}[h]
\centering
\caption{Number of Images in Subsets by Race and Gender}
    \label{table:images}
	\begin{tabular}{l|cccccc|c|cc}
\hline
& \textbf{WF} & \textbf{BF}  & \textbf{WM} & \textbf{BM} & \textbf{dF} & \textbf{dM} &\textbf{Overall} &\textbf{F} & \textbf{M}  \\
\hline
\textbf{S1} & 1,285  & 1,285  & 3855 & 3,855 & 0 &0 & 10,280 & 2570    & 7,710\\
\textbf{S2} & 1,285  & 1,285   & 3,855     & 3,855 &0 & 0 &10,280 & 2,570     & 7,710 \\
\textbf{R}  & 0 & 3,150  & 220    & 28,980 &144 &1,850 & 34,344   & 3,294    &31,050 \\
\hline
\textbf{Overall}  & 2,570 & 5,720  & 7,930 & 36,690  &144 &1,850 & 54,904 & 8,434 & 46,470\\
\hline
\end{tabular}
\newline
Race-gender combinations are abbreviated, e.g., \textbf{BF} represents the black female group. Abbreviations \textbf{dF} and \textbf{dM} represent those who are neither black nor white in race. 
\end{table}

\subsection{Facial Feature Extraction}
In computer vision, image preprocessing is often an essential first step to reduce unnecessary variation, decrease pixel dimension, and simplify pixel encoding. Despite the standard format of police photography in mugshots, Morph-II photographs vary in head-tilt, camera distance, occlusion, and illumination. We address this variation as follows.
Images are first converted to grayscale. 
Next, faces are automatically detected, eliminating the background and hair, so that no external cues can be used to classify gender. The resulting images are centered and scaled with respect to the center of the irises. Finally, the images are cropped to be 70 pixels tall by 60 pixels wide. Full methodological details are given in \cite{Yip2018preprocess} and align with standard preprocessing protocols from face analysis.

After preprocessing, pixel-related features are extracted from the face images in Morph-II. As discussed previously, there are numerous approaches for feature extraction. In this study on Morph-II, we incorporate domain expertise by choosing three well-established appearance-based models from image analysis: local texture techniques such as local binary patterns (LBP) \cite{yang2007demographic,lian2006multi,makinen2008experimental,alexandre2010gender}, biologically-inspired features (BIF) \cite{guo2009gender, han2015demographic}, and histogram of oriented gradients (HOG) \cite{guo2009gender}. Additionally, these model-based approaches provide "robust interpretation $\hdots$ by constraining solutions to be face-like" \cite{edwards1998interpreting}.
Detailed documentation of our feature extraction process can be found in \cite{feature, Yip2018preprocess}. 

\input{Parameter_table.tex}

\subsection{Kernel-Based Dimension Reduction Optimization}
Tuning parameter selection is essential for kernel-based methods in order to achieve good results. Within the framework of feature extraction, dimension reduction, and the classification model, there are many combinations of parameters to be considered. 
The main parameters and tested values are summarized in Table \ref{table:parameter} and discussed as follows.
LBP features have two main parameters: block size $s$ and window radius $r$. For HOG, the two main parameters are block size $s$ and number of orientations $o$. For BIF, we consider the block size $s$ and the parameter $\gamma$, which represents the spatial aspect ratio; there is also a choice of pooling operation, which we select here as the standard deviation operation.

For each dimension reduction method, the radial kernel
\begin{equation}
\label{eq:RBF}
k(x_{i},x_{j})=e^{\delta {||x_{i}-x_{j}||}_2^{2}}
\end{equation}
is used for each pair of observation vectors $x_i,x_j$, based on the results from our simulation studies. 
In the kernel, we must select the tuning parameter $\delta$, which scales the extent of similarity between pairs of vectors. This parameter must be chosen with particular care, since a poor choice can result in transformed features with little to no variability. Empirically, we observed that SKPCA was more sensitive than KLDA and KPCA to the choice of $\delta$; values of $\delta$ at or above $1$ resulted in a rank deficient matrix and failure to compute all requested eigenvalues.
For SKPCA, we consider an additional scaling parameter $\eta$ in the modified link function proposed by Wang et al. \cite{wang2015modified}:
\begin{equation}
\label{eq:link}
l(y_i,y_j)=e^{\eta \delta {||x_{i}-x_{j}||}_2^{2}},
\end{equation}
for all observed responses $y_i,y_j$ in the same class. The scale parameter $\eta$ enables the weighing of dependence between the covariates and response.

Finally, we choose a linear SVM to classify gender based on the dimension-reduced, transformed features. The motivation for this classifier is discussed in the next section. The main parameter for linear SVM is the cost $c$, which measures the extent to which misclassification in training will be permitted. We consider values of $c$ from $10^{-8}$ to $10^{8}$.

\input{Tuning_table.tex}

We tune on small subsets of Morph-II to reduce runtime, memory consumption, and risk of over-fitting. $1000$ images from $S_1$ and $1000$ images from $S_2$ are randomly selected. The standard method of grid search is followed for tuning on these subsets. For each combination of parameters, a model is trained on the subset from $S_1$ and then tested on the subset from $S_2$. 
For each dimension reduction method paired with each feature type (BIF, HOG, and LBP), the best three or four accuracy rates from tuning are obtained. (Except in the case of ties, we choose only the top three accuracy rates.) The tuning results for these top-performing parameters are given in Table \ref{table:tuning}. The parameters corresponding to these maximum accuracy rates are applied to the full dataset through the previously discussed evaluation protocol. Although this protocol involves testing on images from $S_1$ and $S_2$, any overlap of images is minor (in each testing set, less than $2.3\%$ of images have been used in tuning) and has little impact on the reported accuracy (discussed in Section 6). Regardless, the tuning parameters are selected through the same procedure, so the classification performances can be fairly compared among all considered DR methods.


\input{GenderAccuracy_Table.tex}

\subsection{Gender Classification}
For the classification part of the modeling, linear SVM is adopted. Many face analysis studies have involved SVM, as summarized in \cite{byun2002applications}. Briefly, SVM identifies a separating hyperplane with maximal margin between the classes. Several popular kernels for SVM include linear, polynomial, and RBF \cite{steinwart2008support}. We select the linear kernel, because directions of variability in the data are expected to be linear after the nonlinear transformations of KPCA, SKPCA, or KLDA. Indeed, Sch\"olkopf et al. observed this to be true for KPCA in their landmark study \cite{scholkopf1998nonlinear}. The linear kernel for SVM also reduces computational cost, compared to nonlinear kernels.

With the parameters in Table \ref{table:tuning} that are selected from tuning on subsets, we implement dimension reduction and classification on the full Morph-II dataset, following the subsetting scheme discussed in Section 5.2.
The challenges of the large size of Morph-II, the high dimensionality of the features, and the computational complexity of these dimension reduction methods necessitate the use of high-performance computing (HPC). 
For example, the kernel matrix for each dimension reduction method is $55134 \times 55134$, requiring approximately 23 gigabytes of storage. Thus, we implement the process on the HiPerGator 2.0 supercomputing cluster at the University of Florida. The code is written in R. The R package \textit{rARPACK} is used to optimize the solving of eigenvalue problems \cite{qiu2016package}, and the \textit{e1071} package is utilized for training and testing the SVM model \cite{dimitriadou2005misc}.

\section{Experiment Results}
The kernel-based DR methods KPCA, SKPCA, and KLDA are applied to three facial feature extraction methods: BIF, HOG, and LBP. The DR methods transform the feature data, then reduce the dimension. In all cases, a dimension of $100$ is retained, substantially lower than the dimension of the original feature space. The dimensionality of 100 is selected as a trade-off between computation time and classification accuracy based on our preliminary studies. The transformed and dimension-reduced data serve as input for the linear SVM, which classifies each image subject as male or female. Additionally, these predicted gender classes are mapped to probabilities through a sigmoid function, following \cite{platt1999probabilistic}. 
This process is applied to each alternation of the evaluation protocol: 1) train on $S_1$, test on $S_2 \cup R$ and 2) train on $S_2$, test on $S_1 \cup R$. The classification results are averaged over these two testing sets. 
The mean classification accuracy over the testing images is chosen as the evaluation criterion for our methods on Morph-II, as it is the usual performance metric for gender classification \cite{guo2009gender}, especially in similar studies \cite{guo2011simultaneous, guo2014framework, yi2014age, yang2013automatic}.


These mean classification results from Morph-II are shown in Table \ref{table:classify}. 
In addition to the accuracy, the true positive rate (also known as sensitivity or recall) and true negative rate (also called specificity) are given. 
For this study, we define the true positive rate (TPR) as the proportion of male faces correctly classified, while the true negative rate (TNR) as the proportion of female faces correctly classified. 
The memory and runtime are also listed in Table \ref{table:classify}. The runtime is the total time for training and testing on HPC, i.e., the average of time1 (train on $S_1$, test on $S_{2} \cup R$) and time2 (train on $S_2$, test on $S_{1} \cup R$). As mentioned in Section 5.4, there is a small overlap between the tuning and testing sets that could contribute to over-fitting. We have assessed the potential impact of over-fitting on our reported accuracy rates and found it to be very small: it is estimated to be (at most) between 0.09\% and 0.2\% and to monotonically decrease as reported accuracy rates increase.


\begin{figure}
\includegraphics[width=.6\textwidth, trim={0cm 0cm 0cm 2cm},clip]{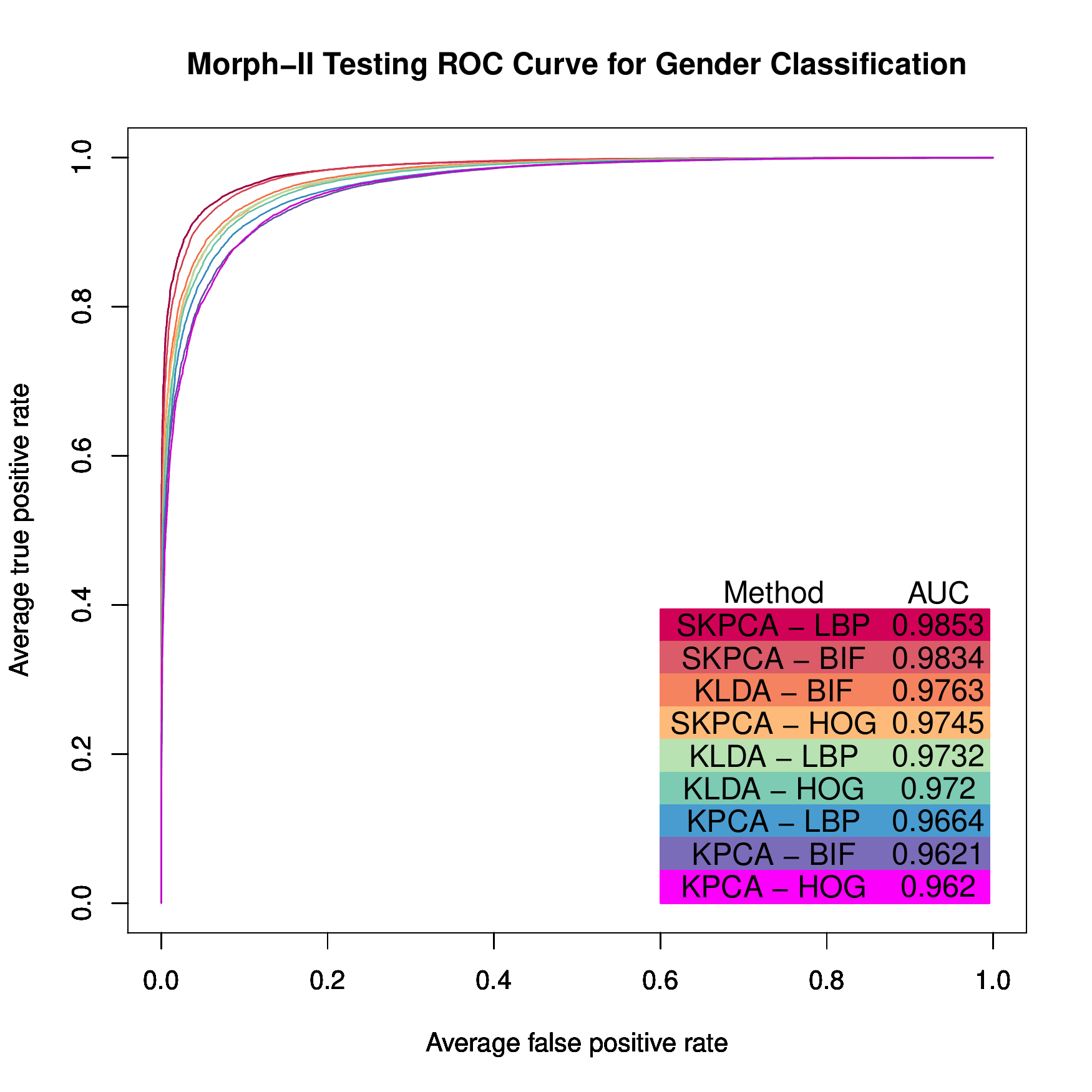}
\centering
\caption{Receiver operating characteristic (ROC) curve and area under the curve (AUC) are compared by method for gender classification on Morph-II. Each color corresponds to a DR method paired with feature type. For each probability threshold, the true and false positive rates are reported as the averages from the testing sets of the alternating evaluation protocol.}
\label{fig:Morph_ROC}
\end{figure}

The classification performance is further visualized in Figure \ref{fig:Morph_ROC} through receiver operating characteristic (ROC) curves for the nine combinations of DR method and feature extraction type. For each combination, its displayed curve corresponds to the "best" results from Table \ref{table:classify} (the combination of parameters reaching maximum mean classification accuracy or maximum mean true positive rate in the event of ties). 
For each alternation of the evaluation protocol, the true and false positive rates in testing are calculated for each probability threshold. To construct the ROC curves, each of the resulting rates for each threshold is averaged over the testing sets. 

Table \ref{table:classify} shows that for the feature BIF, SKPCA and KLDA outperform KPCA. For the feature HOG, SKPCA achieves higher accuracy than both KPCA and KLDA, while the latter two techniques perform very similarly. Last, for the feature LBP, SKPCA produces better classification accuracy than KPCA and KLDA. In summary, our experiment's results indicate that SKPCA outperforms KLDA consistently, while KLDA outperforms KPCA for all three features BIF, LBP, and HOG. 
On the other hand, for KPCA, the features HOG and LBP produce approximately the same accuracies, outperforming BIF. For SKPCA, LBP achieves slightly better results than BIF, while LBP and BIF both outperform HOG. Finally, for KLDA, BIF reaches slightly higher accuracy than LBP, while BIF and LBP both exceed HOG.

In most cases, the accuracy (in Table \ref{table:classify}) and AUC (in Figure \ref{fig:Morph_ROC}) metrics agree on the best methods. An exception is that SKPCA with the HOG features achieves slightly higher accuracy ($94.89\%$) than KLDA with the BIF features ($94.18\%$), but SKPCA with HOG has lower AUC than KLDA with BIF. The other exception is that KPCA with the HOG features has the lowest AUC of the nine methods, but its accuracy is higher than KPCA with the BIF features. In summary, the accuracy and AUC results imply that SKPCA generally performs best for gender classification on Morph-II, while KLDA tends to outperform KPCA. Meanwhile, the LBP and BIF features often yield better classification performance, with less memory usage, than the HOG features. 

It is interesting that, overall, LBP achieves even slightly better performance than BIF for the dimension reduction method SKPCA on the task of gender classification, since BIF is popular in demographic analysis such as age estimation, gender classification, and race classification \cite{guo2009gender,guo2010study,guo2010human,guo2011simultaneous,guo2014framework}. Another interesting fact is displayed by the results of the true positive and negative rates in Table \ref{table:classify}: males have a higher probability of correct identification than females, with the biggest margin exceeding 20\%. Our finding is consistent with \cite{han2015demographic}: females are more challenging to correctly classify than males, both for automatic approaches and human perception. Similarly, for race classification on Morph-II, Guo and Mu found in \cite{guo2010study} that training a model on female faces (and testing on male faces) contributed to significantly more errors on average than training on male faces (and testing on female faces), even when controlling for differences in the training sample sizes. Our results also indicate that, overall, HOG and LBP outperform BIF for males, while BIF works consistently better than LBP and HOG for females. 

Next, in Table \ref{comparison} we compare our results to studies using similar methods on Morph-II, as well as recent state-of-the-art works with deep learning on MORPH-II. With the exception of \cite{han2015demographic}, all studies' results in the table are mean testing classification accuracy from an alternating evaluation protocol based on Guo et al \cite{guo2010human}. Hence, our results can be directly compared to these studies. With LBP features, SKPCA, and a linear support vector machine (SVM), our gender classification accuracies approximate 96\%, competitive with benchmark results. Interestingly, several reported accuracy rates from human observers of gender range from 96\% \cite{burton1993s} to 96.9\%  \cite{han2015demographic}.
The similarity in recognition rates between our methods and human observers can further validate the success of our approach.

\input{Comparison_table.tex}

\section{Kernel-based Dimension Reduction Optimization and Classification on FG-NET}
For further comparison between KPCA, SKPCA, and KLDA, we apply a modification of our approach from Section 5 to a smaller face dataset, the face and gesture recognition network (FG-NET).
FG-NET is a popular, publicly available database used for age estimation, gender classification, face recognition, and other demographic analysis tasks \cite{panis2016overview}. It contains 1002 images from 82 subjects: 47 males and 35 females with ages varying from 0 to 69 years \cite{panis2016overview}. 

For each image, 109 features are extracted using the Active Appearance Model (AAM), a commonly adopted appearance-based approach that models the shape and texture of the face \cite{edwards1998interpreting, cootes1998active}.
As in Section 5.4, the radial kernel defined in equation (\ref{eq:RBF}) is chosen for each of the DR methods KPCA, SKPCA, and KLDA. Additionally, the modified link function from equation (\ref{eq:link}) is applied in the SKPCA algorithm. Thus, the tuning parameter $\delta$ in the radial kernel and $\eta$ in the modified link function must be selected. As in our experiments on Morph-II, linear SVM is chosen as the classifier for FG-NET. On Morph-II, values of the cost parameter $c$ ranging from $10^{-8}$ to $10^8$ were tested. On FG-NET, we have observed convergence issues in the SVM algorithm for values of $c$ exceeding 10, so only the values $10^{-8},10^{-7},\hdots,10^{-1},1,10$ are tested. The considered tuning parameters are summarized in Table \ref{table:parameter_FG}.

\renewcommand{\arraystretch}{1.3}

\input{Parameter_table_FG.tex}

\renewcommand{\arraystretch}{1}

For cross-validation, we use leave-one-person-out (LOPO), the most well-accepted scheme for FG-NET \cite{panis2016overview}. LOPO is a variation of $k$-fold cross-validation that produces independent training and testing folds in longitudinal datasets. The number of folds $k$ is set equal to the number of subjects in the dataset, so $k=82$ here. For $i=1,2,\hdots 82$, testing fold $i$ contains only images of person $i$, while training fold $i$ contains all remaining images. 
Similarly to on Morph-II, we choose the mean classification accuracy over the testing folds to be the evaluation criterion.

\input{Accuracy_Table_FG.tex}

\begin{figure}
\includegraphics[width=.6\textwidth]{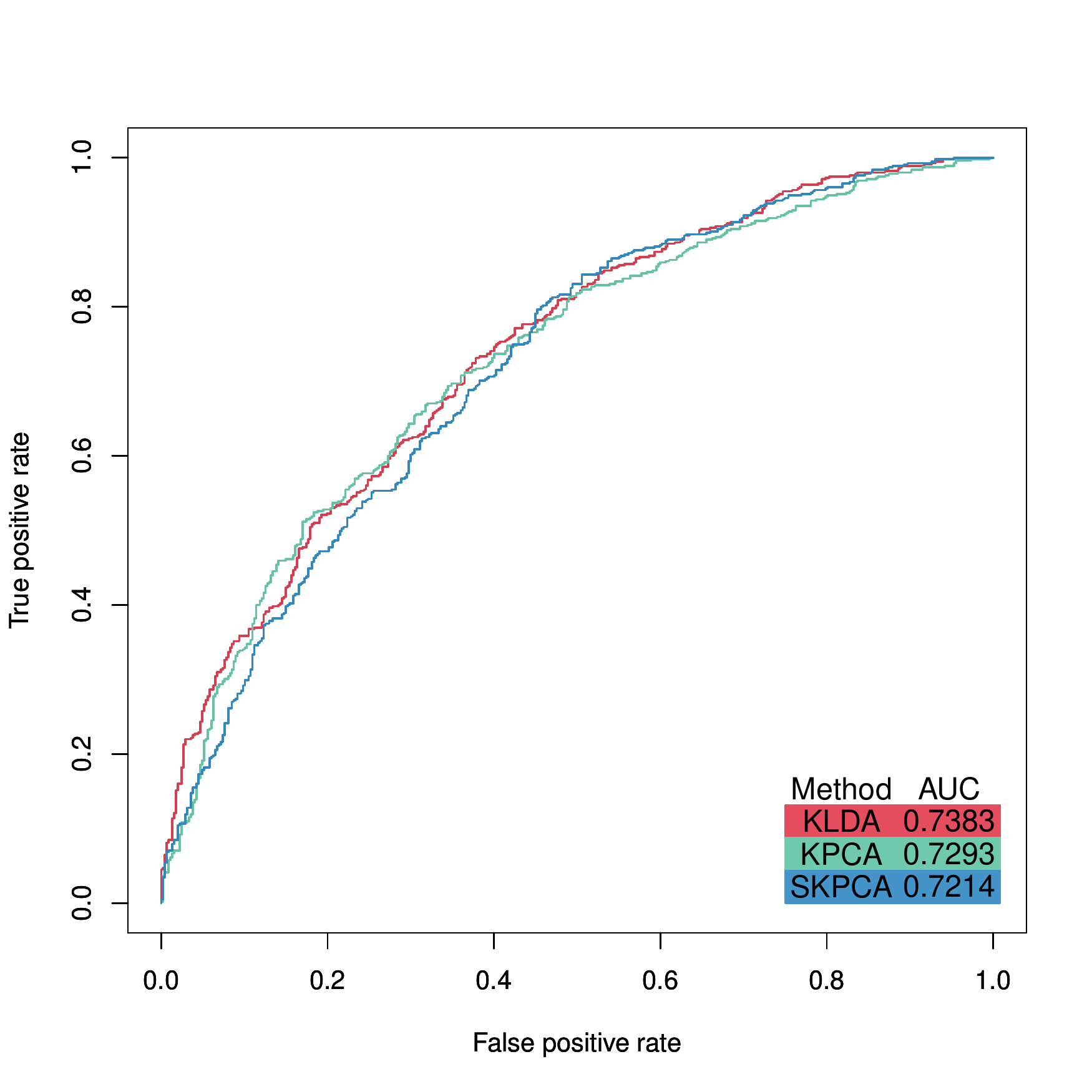}
\centering
\caption{
Receiver operating characteristic (ROC) curve and area under the curve (AUC) are compared by method for gender classification on FG-NET. Each color corresponds to a DR method.}
\label{fig:FG_ROC}
\end{figure}

For each fold, we transform and reduce the dimension of the features through each DR method. In all cases, a dimension of 100 is retained to facilitate comparison with the results on Morph-II. The transformed, dimension-reduced features then predict the gender of the testing fold's images through a linear SVM. The predicted classes from SVM are also mapped to probabilities through \cite{platt1999probabilistic}, similarly as in Section 6. The gender classification accuracy is calculated for the testing fold. Finally, all such testing classification accuracies are averaged to compute the mean classification accuracy from testing; the testing probabilities are used to form ROC curves.  

The optimum gender classification results on FG-NET are presented in Table \ref{table:classify_FG}. The maximum classification accuracy of about 72.25\% is achieved by KLDA. For other choices of parameters, KLDA reaches above 71\% accuracy, which is close to the maximum accuracy attained by SKPCA. Meanwhile, the peak accuracy reached by KPCA is 70.25\%. In general here, KLDA is observed to outperform SKPCA and KPCA, while SKPCA tends to surpass KPCA. In most cases, the probability of correctly classifying males (sensitivity/true positive rate) is higher than the probability of correctly classifying females (specificity/true negative rate). For each DR method, an ROC curve (corresponding to the results from Table \ref{table:classify_FG} with maximal mean classification accuracy) is displayed in Figure \ref{fig:FG_ROC}. The area under the curve (AUC) is highest for KLDA, followed by KPCA then SKPCA. 

Overall, the gender classification results on Morph-II are stronger than on FG-NET. Lower accuracy on FG-NET could be caused by the greater number of minors (aged 0-18), who have been more difficult to classify than adults in some studies \cite{wang2015modified, wang2010gender}.
Additionally, there are substantially fewer faces for training in FG-NET versus Morph-II (under 1000 versus 10280 images). Another contributor could be the choice of features and its dimension; the AAM features have dimension 109 on FG-NET, while the HOG, LBP, and BIF features have dimensions ranging from 500 to thousands on Morph-II. SKPCA reaches peak performance on Morph-II, while KLDA attains optimal results on FG-NET. However, the results on Morph-II and FG-NET are similar in that the supervised methods KLDA and SKPCA outperform the unsupervised method KPCA for gender classification. Further, both datasets evidence that female faces are more challenging to classify than male faces. 

\section{Computational Framework for Practical Systems}
To tackle the challenges of high dimensionality and intensive computation for large-scale databases (like Morph-II, as shown in  the Time column of Table \ref{table:classify}) in real-world applications, we propose a computational framework to substantially decrease runtime.

\begin{figure}
\includegraphics[width=0.9\textwidth]{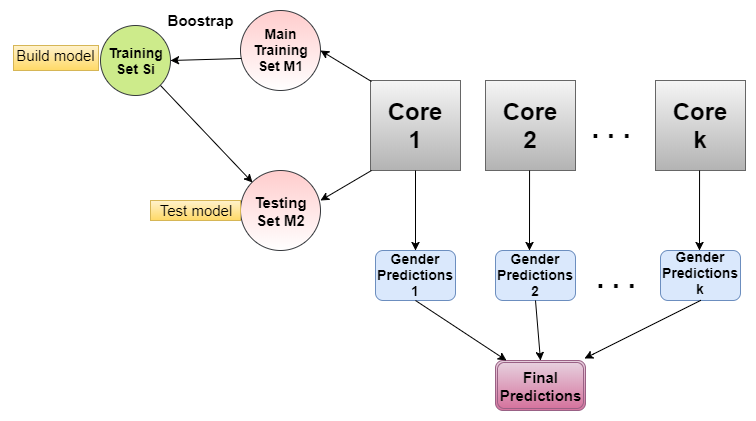}
\centering
\caption{Flowchart representing the parallel computational framework for practical systems proposed for Morph-II and other large datasets.}
\label{fig:flowchart}
\end{figure}

Our approach involves parallel computing, the bootstrap resampling method, and ensemble learning. Let $M1$ denote the main training set and $M2$ the testing set. If $M1$ is very large, we can save some time by drawing bootstrapped samples from $M1$. Let $S_i$ denote the $i$th bootstrapped sample from $M1$. Send $S_i$ to a core (or processor), Core $i$. Train the model on $S_i$. Test on the full testing set $M2$, obtaining a set of gender predictions corresponding to Core $i$ and $S_i$. Repeat this process for all bootstrapped samples and corresponding cores $i$. The final predictions are obtained by taking the majority rule of the predictions from all $i$ cores and samples. Hence, the results from this scheme approximate the results from the full Morph-II. This framework is summarized in Figure \ref{fig:flowchart}.

To explore the effectiveness, this framework is applied to Morph-II with a selection of BIF, LBP, and HOG features as preliminary studies. This experiment is implemented through the
HiPerGator 2.0 supercomputer at University of Florida with five cores per combination of feature and dimension reduction method. Following the subsetting scheme discussed in Section 5.2, for simplicity, we consider only the case of bootstrapping image samples from $S_1$ for training, while each image from  $S_2 \cup R$ is used for testing.

\begin{table}[ht]
\centering
\caption{Classification Results Based on Bootstrapping}
\label{bootstrap}
\small
\begin{tabular}{|c|l|l|l|l|}
\hline
Method &Feature &Accuracy &Memory (gb) &Time (min) \\
\hline
\multirow{2}{*}{\begin{tabular}[c]{@{}c@{}}
\textbf{KPCA}\\ \end{tabular}}
& \begin{tabular}[c]{@{}c@{}}BIF $s=s7-37,\gamma=0.4$ \end{tabular} &\textbf{0.9330} &27.59 &90\\
& \begin{tabular}[c]{@{}c@{}}HOG $s=12,o=8$  \end{tabular} &0.9178  &29.77 &101 \\
& \begin{tabular}[c]{@{}c@{}}LBP $s=10,r=1$ \end{tabular}  &0.8927 &25.85 &37 \\
\hline
\multirow{2}{*}{\begin{tabular}[c]{@{}c@{}}\textbf{SKPCA} \end{tabular}}
& \begin{tabular}[c]{@{}c@{}}BIF $s=s7-37,\gamma=0.4$ \end{tabular} &\textbf{0.9417} &53.28 &89\\
& \begin{tabular}[c]{@{}c@{}}HOG $s=12,o=8$ \end{tabular} &0.9056 &51.43 &74\\
& \begin{tabular}[c]{@{}c@{}}LBP $s=10,r=1$ \end{tabular} &0.9274 &20.33 &24
\\
\hline
\multirow{2}{*}{\begin{tabular}[c]{@{}c@{}}\textbf{KLDA} \end{tabular}}
& \begin{tabular}[c]{@{}c@{}}BIF $s=s7-37,\gamma=0.4$ \end{tabular} &\textbf{0.9416} &30.99 &100\\
& \begin{tabular}[c]{@{}c@{}}HOG $s=12,o=8$ \end{tabular} &0.9133 &25.42 &102\\
& \begin{tabular}[c]{@{}c@{}}LBP $s=10,r=1$ \end{tabular} &0.9118 &17.05 &26\\
\hline
\end{tabular}
\end{table}
\normalsize

We evaluate this framework by comparing the approximated results in Table \ref{bootstrap} to the results from Table \ref{table:classify}. For each combination of feature and dimension reduction method, each of the five cores independently trains a bootstrapped sample of $1000$ images from $S_1$ and tests on $S_2 \cup R$. Then the gender predictions over all five cores are compared with a simple majority rule; e.g., if an image is predicted male for three images and female for two images, the final gender prediction is male. The times in Table \ref{bootstrap} are the total runtimes for this process, which include training and testing on HPC. Therefore, the times and memory can be compared between Tables \ref{table:classify} and \ref{bootstrap}. A distinction is that in Table \ref{table:classify}, results are averaged for the alternating scheme, while in Table \ref{bootstrap}, the results are only from when $S_1$ is used for training and $S_2 \cup R$ for testing.

It is shown in Table \ref{bootstrap} that, in many cases, the accuracy rates from the approximations are similar to those from the main approach in Table \ref{table:classify}. This is a very good result, especially considering that the bootstrapping approach uses no more than $5000$ images total for training, while the main approach used all $10280$ images for training. This finding suggests that our methods may perform reasonably well on Morph-II with smaller training sets. The most substantial difference between the bootstrapped approach and the main approach is in the runtime. For all combinations of features and dimension reduction methods, the bootstrapping approach has decreased the runtime to under two hours. Meanwhile, the main approach in Table \ref{table:classify} yields runtimes exceeding 20 hours. Hence, our preliminary results indicate the parallel approximation approach can attain similar accuracy rates to the main approach, while substantially saving time. Such a result is promising for practical gender classification systems, where gender predictions must be made in real-time.

\section{Conclusion}
We have performed a comparative study of the nonlinear dimension reduction methods KPCA, SKPCA, and KLDA. These kernel-based methods are first applied to three simulated datasets for visualization and comparison. SKPCA and KLDA outperform KPCA, reinforcing the need for supervised approaches in classification tasks. The radial kernel performed well, encouraging its use for face analysis.

Next, we have proposed and evaluated a new machine learning process for Morph-II. First, we use a novel subsetting scheme that reduces class imbalances while establishing independence between training and testing sets. Then we preprocess Morph-II photographs and extract three appearance-based features: HOG, LBP, and BIF. We transform and reduce the dimension of these features through KPCA, SKPCA, and KLDA. Linear SVM classifies the gender of Morph-II subjects, reaching accuracy rates of 95\%. With promising preliminary results on Morph-II, a practical computational framework is offered that reduces runtime through parallelization and approximation.

The performance of the dimension reduction methods are further compared through an application to the FG-NET dataset. Images are represented through the appearance-based AAM features; transformed and reduced in dimension through KPCA, SKPCA, and KLDA; and classified as containing a male or female subject through linear SVM. While SKPCA performed optimally on Morph-II, KLDA reached top performance on FG-NET with 72\% leave-one-person-out (LOPO) accuracy.

Further directions of research involve automatic tuning parameter selection, reduction of computational cost, and application to other face analysis tasks. Our approach could yield improved results with better choices of parameters, but it is impossible to anticipate and try all combinations. Automatic parameter selection for kernels could help identify a good set of parameters more easily. 
Perhaps the most important future direction of research on Morph-II is to reduce computational cost. For many practical demographic analysis systems, predictions must be made in real-time. For our gender classification methods, our parallel approximation approach substantially reduced runtime while attaining similar accuracy rates to the main approach. Such computational strategies should be further investigated to help bring gender classification and other face analysis tasks to practical implementation. Finally, our machine learning pipeline for Morph-II could be generalized to race classification or even age estimation. 


\section{Acknowledgments}
\noindent
This material is based in part upon work supported by the National Science Foundation under Grant Numbers DMS-1659288. Any opinions, findings, and conclusions or recommendations expressed in this material are those of the author(s) and do not necessarily reflect the views of the National Science Foundation. The authors would like to thank the reviewers for the helpful comments that significantly improves the presentation of the paper.

\bibliographystyle{plain}
\bibliography{bibliography}

\end{document}

%% file: Parameter_table.tex
\begin{table}[h]
\centering
\caption{Parameter Summary}
\label{table:parameter}
\begin{tabular}{|c|c|l|}
\hline
\multirow{3}{*}{\begin{tabular}[c]{@{}c@{}}
\textbf{Features}\\ \end{tabular}}
&LBP &$s=10,12,14,16,18,20$\\ & &$r=1,2,3$  \\ \cline{2-3}
& \begin{tabular}[c]{@{}c@{}}HOG\end{tabular} &$s=4,6,8,10,12,14$\\ & &$o=4,6,8$ \\ \cline{2-3}
& \begin{tabular}[c]{@{}c@{}}BIF\end{tabular} &s=$7-37,15-29$\\& &$\gamma=0.1,0.2,\hdots,1.0$  \\
\hline
\multirow{3}{*}{\begin{tabular}[c]{@{}c@{}}\textbf{Dimension Reduction} \end{tabular}}
& KPCA &$\delta=\pm 0.1,\pm 1,\pm 5,\pm 10, \pm 100$  \\ \cline{2-3}
& \begin{tabular}[c]{@{}c@{}}SKPCA \end{tabular} & $\delta=$-0.0001,-0.001 \\ & &$\eta=0.001,0.01,0.1,1$ \\ \cline{2-3}
 & \begin{tabular}[c]{@{}c@{}}KLDA \end{tabular} & $\delta=\pm 0.01,\pm 0.1,\pm 1,\pm 5,\pm 10,\pm 100$ \\ \hline
\textbf{Classifier} &Linear SVM & $c=10^{-8},\hdots,10^{-1},1,10,\hdots,10^8$\\
\hline
\end{tabular}
\end{table}

%% file: Tuning_table.tex
\begin{table}[h]
\centering
\caption{Tuning Results on a Subset of MORPH-II}
\label{table:tuning}
\small
\begin{tabular}{|c|l|l|c|}
\hline
Method&Feature &Parameters &Accuracy\\
\hline
\multirow{3}{*}{\begin{tabular}[c]{@{}c@{}}
\textbf{KPCA}\\ \end{tabular}}
& \begin{tabular}[c]{@{}c@{}}BIF $s=7-37,\gamma=0.1$ \end{tabular} &$\delta=-1,c=10$ &0.882\\
& \begin{tabular}[c]{@{}c@{}}BIF $s=7-37,\gamma=0.6$ \end{tabular} &$\delta=-1,c=10$ &0.882  \\
& \begin{tabular}[c]{@{}c@{}}BIF $s=15-29,\gamma=0.1$ \end{tabular} &$\delta=-1,c=100$ &0.882\\
& \begin{tabular}[c]{@{}c@{}}BIF $s=15-29,\gamma=0.6$ \end{tabular} &$\delta=-1,c=10$ &0.882 \\
\cline{2-4}
& \begin{tabular}[c]{@{}c@{}}HOG $s=4,o=4$ \end{tabular} &$\delta=-100,c=0.1$ &0.917  \\
& \begin{tabular}[c]{@{}c@{}}HOG $s=4,o=4$ \end{tabular} &$\delta=-5,c=0.001$ &0.919 \\
& \begin{tabular}[c]{@{}c@{}}HOG $s=4,o=4$ \end{tabular} &$\delta=-1,c=0.001$ &0.917   \\
& \begin{tabular}[c]{@{}c@{}}HOG $s=4,o=4$ \end{tabular} &$\delta=-0.1,c=0.1$ &0.917  \\
\cline{2-4}
& \begin{tabular}[c]{@{}c@{}}LBP $s=10,r=1$ \end{tabular} &$\delta=-100,c=0.1$ &0.912 \\
& \begin{tabular}[c]{@{}c@{}}LBP $s=10,r=1$ \end{tabular} &$\delta=-5,c=0.1$ &0.912 \\
& \begin{tabular}[c]{@{}c@{}}LBP $s=10,r=1$ \end{tabular} &$\delta=-1,c=0.001$ &0.912  \\
& \begin{tabular}[c]{@{}c@{}}LBP $s=10,r=1$ \end{tabular} &$\delta=-0.1,c=0.1$ &0.912 \\
\hline
\multirow{3}{*}{\begin{tabular}[c]{@{}c@{}}\textbf{SKPCA} \end{tabular}}
& \begin{tabular}[c]{@{}c@{}}BIF $s=7-37,\gamma=0.2$ \end{tabular} &$\delta=0.0001,\eta=0.1,c=1$ &0.899 \\
& \begin{tabular}[c]{@{}c@{}}BIF $s=7-37,\gamma=0.8$ \end{tabular} &$\delta=0.0001,\eta=0.1,c=1$ &0.899 \\
& \begin{tabular}[c]{@{}c@{}}BIF $s=15-29,\gamma=0.5$ \end{tabular} &$\delta=0.0001,\eta=0.1,c=1$ &0.899 \\
\cline{2-4}
& \begin{tabular}[c]{@{}c@{}}HOG $s=6,o=6$ \end{tabular} &$\delta=0.0001,\eta=0.001,c=1$ &0.931\\
& \begin{tabular}[c]{@{}c@{}}HOG $s=6,o=6$ \end{tabular} &$\delta=0.0001,\eta=0.01,c=0.001$ &0.931 \\
& \begin{tabular}[c]{@{}c@{}}HOG $s=6,o=6$ \end{tabular} &$\delta=0.0001,\eta=0.1,c=0.001$ &0.931
\\
\cline{2-4}
& \begin{tabular}[c]{@{}c@{}}LBP $s=14,r=2$ \end{tabular} &$\delta=0.0001,\eta=0.001,c=1$ &0.937
\\
& \begin{tabular}[c]{@{}c@{}}LBP $s=14,r=2$ \end{tabular} &$\delta=0.0001,\eta=0.01,c=1$ &0.937
\\
& \begin{tabular}[c]{@{}c@{}}LBP $s=14,r=2$ \end{tabular} &$\delta=0.0001,\eta=0.1,c=1$ &0.938\\
& \begin{tabular}[c]{@{}c@{}}  LBP  $s=14,r=2$ \end{tabular} &  $\delta=0.0001,\eta=1,c=1$ & 0.939
\\
\hline
\multirow{3}{*}{\begin{tabular}[c]{@{}c@{}}\textbf{KLDA} \end{tabular}}
& \begin{tabular}[c]{@{}c@{}}BIF $s=7-37,\gamma=0.3$ \end{tabular} &$\delta=-1,c=10$ &0.875 \\
& \begin{tabular}[c]{@{}c@{}}BIF $s=7-37,\gamma=0.6$ \end{tabular} &$\delta=-1,c=100$ &0.875\\
& \begin{tabular}[c]{@{}c@{}}BIF $s=15-29,\gamma=0.2$ \end{tabular} &$\delta=-1,c=10$ &0.875\\
& \begin{tabular}[c]{@{}c@{}}BIF $s=15-29,\gamma=0.8$ \end{tabular} &$\delta=-1,c=100$ &0.875\\
\cline{2-4}
& \begin{tabular}[c]{@{}c@{}}HOG $s=4,o=4$ \end{tabular} &$\delta=1,c=1$ &0.917 \\
& \begin{tabular}[c]{@{}c@{}}HOG $s=4,o=6$ \end{tabular} &$\delta=1,c=1$ &0.917 \\
& \begin{tabular}[c]{@{}c@{}}HOG $s=12,o=8$ \end{tabular} &$\delta=-1,c=100$ &0.904 \\
\cline{2-4}
& \begin{tabular}[c]{@{}c@{}}LBP $s=10,r=1$ \end{tabular} &$\delta=-0.1,c=1$ &0.906 \\
& \begin{tabular}[c]{@{}c@{}}LBP $s=10,r=1$ \end{tabular} &$\delta=1,c=1$ &0.908  \\
& \begin{tabular}[c]{@{}c@{}}LBP $s=14,r=1$ \end{tabular} &$\delta=0.1,c=10$ &0.898  \\
\hline
\end{tabular}
\end{table}
\normalsize

%% file: GenderAccuracy_Table.tex
\begin{table}[h]
\caption{Gender Classification Results on MORPH-II}
\label{table:classify}
\small
\begin{adjustbox}{width=0.99\textwidth}
\begin{tabular}{|c|l|l|l|l|l|l|l|}
\hline
Method&Feature &Parameters &Acc\textsuperscript{(1)}
&TPR\textsuperscript{(2)} &TNR\textsuperscript{(3)} &Mem\textsuperscript{(4)} &Time\textsuperscript{(5)}\\
\hline
\multirow{3}{*}{\begin{tabular}[c]{@{}c@{}}
\textbf{KPCA}\\ \end{tabular}}
& \begin{tabular}[c]{@{}c@{}}BIF $s=7-37,\gamma=0.1$ \end{tabular} &$\delta=-1,c=10$ &0.9296 &0.9473 &0.8127 &34.04 &42.26 \\
& \begin{tabular}[c]{@{}c@{}}BIF $s=7-37,\gamma=0.6$ \end{tabular} &$\delta=-1,c=10$ &\textbf{0.9297} &0.9455 &0.8112 &34.68 &36.94 \\
& \begin{tabular}[c]{@{}c@{}}BIF $s=15-29,\gamma=0.1$ \end{tabular} &$\delta=-1,c=100$ &0.9071 &0.9377 &0.7050 &31.74 &33.83 \\
& \begin{tabular}[c]{@{}c@{}}BIF $s=15-29,\gamma=0.6$ \end{tabular} &$\delta=-1,c=10$ &0.9096 &0.9374 &0.7266 &31.80 &35.97 \\
\cline{2-8}
& \begin{tabular}[c]{@{}c@{}}HOG $s=4,o=4$ \end{tabular} &$\delta=-100,c=0.1$ &\textbf{0.9391} &0.9726 &0.7172 &34.00 &31.54\\
& \begin{tabular}[c]{@{}c@{}}HOG $s=4,o=4$ \end{tabular} &$\delta=-5,c=0.001$ &\textbf{0.9391} &0.9727 &0.7170 &34.00 &30.86\\
& \begin{tabular}[c]{@{}c@{}}HOG $s=4,o=4$ \end{tabular} &$\delta=-1,c=0.001$ &\textbf{0.9391} &0.9724 &0.7192 &34.00 &32.17\\
& \begin{tabular}[c]{@{}c@{}}HOG $s=4,o=4$ \end{tabular} &$\delta=-0.1,c=0.1$ &0.9364 &0.9626 &0.7634&34.35 &31.41\\
\cline{2-8}
& \begin{tabular}[c]{@{}c@{}}LBP $s=10,r=1$ \end{tabular} &$\delta=-100,c=0.1$ &\textbf{0.9391} &0.9726 &0.7172 &34.00 &31.54\\
& \begin{tabular}[c]{@{}c@{}}LBP $s=10,r=1$ \end{tabular} &$\delta=-5,c=0.1$ &\textbf{0.9391} &0.9726 &0.7172 &34.00 &30.86\\
& \begin{tabular}[c]{@{}c@{}}LBP $s=10,r=1$ \end{tabular} &$\delta=-1,c=0.001$ &\textbf{0.9391} &0.9724 &0.7192 &34.00 &32.17\\
& \begin{tabular}[c]{@{}c@{}}LBP $s=10,r=1$ \end{tabular} &$\delta=-0.1,c=0.1$ &0.9364 &0.9626 &0.7634 &34.35 &31.41\\
\hline
\multirow{3}{*}{\begin{tabular}[c]{@{}c@{}}\textbf{SKPCA} \end{tabular}}
& \begin{tabular}[c]{@{}c@{}}BIF $s=7-37,\gamma=0.2$ \end{tabular} &$\delta=0.0001,\eta=0.1,c=1$ &0.9507 &0.9616 &0.8781 &35.48 &42.04 \\
& \begin{tabular}[c]{@{}c@{}}BIF $s=7-37,\gamma=0.8$ \end{tabular} &$\delta=0.0001,\eta=0.1,c=1$ &\textbf{0.9532} &0.9639 &0.8823 &33.04 &38.34 \\
& \begin{tabular}[c]{@{}c@{}}BIF $s=15-29,\gamma=0.5$ \end{tabular} &$\delta=0.0001,\eta=0.1,c=1$ &0.9260 &0.9477 &0.7827 &20.03 &34.58 \\
\cline{2-8}
& \begin{tabular}[c]{@{}c@{}}HOG $s=6,o=6$ \end{tabular} &$\delta=0.0001,\eta=0.001,c=1$ &0.9467 &0.9645 &0.8292 &36.69 &37.39 \\
& \begin{tabular}[c]{@{}c@{}}HOG $s=6,o=6$ \end{tabular} &$\delta=0.0001,\eta=0.01,c=0.001$ &\textbf{0.9489} & \textbf{0.9786} &0.7528 &38.28 &53.96 \\
& \begin{tabular}[c]{@{}c@{}}HOG $s=6,o=6$ \end{tabular} &$\delta=0.0001,\eta=0.1,c=0.001$ &0.9488 &\textbf{0.9786} &0.7517 &39.83 &60.55 \\
\cline{2-8}
& \begin{tabular}[c]{@{}c@{}}LBP $s=14,r=2$ \end{tabular} &$\delta=0.0001,\eta=0.001,c=1$ &\textbf{0.9585}&0.9727 &0.8641 &28.68 &25.33 \\
& \begin{tabular}[c]{@{}c@{}}LBP $s=14,r=2$ \end{tabular} &$\delta=0.0001,\eta=0.01,c=1$ &\textbf{0.9585} &0.9764 &0.8642 &23.22 &38.42 \\
& \begin{tabular}[c]{@{}c@{}}LBP $s=14,r=2$ \end{tabular} &$\delta=0.0001,\eta=.1,c=1$ &\textbf{0.9585} &0.9730 &0.8640 &29.87 &28.00 \\
& \begin{tabular}[c]{@{}c@{}}LBP $s=14,r=2$ \end{tabular} &$\delta=0.0001,\eta=1,c=1$ &\textbf{0.9585} &0.9727 &0.8640 &27.92 &22.83 \\
\hline
\multirow{3}{*}{\begin{tabular}[c]{@{}c@{}}\textbf{KLDA} \end{tabular}}
& \begin{tabular}[c]{@{}c@{}}BIF $s=7-37,\gamma=0.3$ \end{tabular} &$\delta=-1,c=10$ &0.9415 &0.9539 &0.8594 &24.89 &34.50 \\
& \begin{tabular}[c]{@{}c@{}}BIF $s=7-37,\gamma=0.6$ \end{tabular} &$\delta=-1,c=100$ &\textbf{0.9426} &0.9558 &\textbf{0.8858} &24.74 &35.46 \\
& \begin{tabular}[c]{@{}c@{}}BIF $s=15-29,\gamma=0.2$ \end{tabular} &$\delta=-1,c=10$ &0.9131 &0.9374 &0.7532 &22.80 &26.78 \\
& \begin{tabular}[c]{@{}c@{}}BIF $s=15-29,\gamma=0.8$ \end{tabular} &$\delta=-1,c=100$ &0.9205 &0.9421 &0.7783 &22.83 &33.88 \\
\cline{2-8}
& \begin{tabular}[c]{@{}c@{}}HOG $s=4,o=4$ \end{tabular} &$\delta=1,c=1$ &0.9369 &0.9517 &0.8392 &36.52 &81.71 \\
& \begin{tabular}[c]{@{}c@{}}HOG $s=4,o=6$ \end{tabular} &$\delta=1,c=1$ &\textbf{0.9398} &0.9545 &0.8425 &52.48 &148.24 \\
& \begin{tabular}[c]{@{}c@{}}HOG $s=12,o=8$ \end{tabular} &$\delta=-1,c=100$ &0.9175 &0.9421 &0.7542 &21.18 &21.57 \\
\cline{2-8}
& \begin{tabular}[c]{@{}c@{}}LBP $s=10,r=1$ \end{tabular} &$\delta=-0.1,c=1$ &\textbf{0.9418} &0.9578 &0.8428 &24.58 &37.17 \\
& \begin{tabular}[c]{@{}c@{}}LBP $s=10,r=1$ \end{tabular} &$\delta=1,c=1$ &0.9417 &0.9558 &0.8486 &24.70 &36.45 \\
& \begin{tabular}[c]{@{}c@{}}LBP $s=14,r=1$ \end{tabular} &$\delta=0.1,c=10$ &0.9392 &0.9543 &0.8397 &20.77 &31.12 \\
\hline
\addlinespace[1ex]
\end{tabular}
\end{adjustbox}
\footnotesize 
(1) Acc represents mean accuracy. \\
(2) TPR represents mean true positive rate (recall/sensitivity): the proportion of male faces correctly classified. \\
(3) TNR represents mean true negative rate (specificity): the proportion of female faces correctly classified. \\
(4) Mem represents mean memory in gigabytes. \\
(5) Time represents mean runtime in hours for training and testing. \\
\end{table}
\normalsize

%% file: Comparison_table.tex
\begin{table}[H]
\caption{Comparison Results  for Gender Classification on MORPH-II}
\centering
\begin{tabular} {|l|l|l|l|}
\hline
Method & Accuracy & Reference &Year\\
\hline
BIF+OLPP & 98\% & \cite{guo2011simultaneous} & 2011 \\ 
BIF+PLS & 97.34\% &\cite{guo2011simultaneous} & 2011\\
BIF+KPLS & 98.2\% &\cite{guo2011simultaneous} & 2011\\
BIF+CCA & 95.2\% & \cite{guo2014framework} & 2014 \\
BIF+KCCA & 98.4\% & \cite{guo2014framework} & 2014 \\
BIF+rCCA & 97.6\% & \cite{guo2014framework} & 2014 \\
Multi-scale CNN & 97.9\%  & \cite{yi2014age}& 2014\\
Ranking CNN & 97.9\% &  \cite{yang2013automatic} & 2015\\
BIF+Hierarchical-SVM & 97.6\% & 
\cite{han2015demographic} & 2015\\
Human Estimators & 96.9\% & 
\cite{han2015demographic} & 2015\\
LBP+SKPCA+L-SVM & 95.85\% &  This work & 2019\\ 
\hline
\end{tabular}
\label{comparison}
\end{table}

%% file: Parameter_table_FG.tex
\begin{table}[h]
\centering
\caption{Parameter Summary for FG-NET}
\label{table:parameter_FG}
\begin{tabular}{|c|c|l|}
\hline
\multirow{3}{*}{\begin{tabular}[c]{@{}c@{}}\textbf{Dimension Reduction} \end{tabular}}
& KPCA &$\delta=3.2, 3.2\overline{6},3.\overline{3},3.4,3.4\overline{6},3.5\overline{3},3.6,3.\overline{6},3.7\overline{3},3.8$ \\ \cline{2-3}
& \begin{tabular}[c]{@{}c@{}}SKPCA \end{tabular} & $\delta=0.0098$ \\ & &$\eta=0.001,0.01,0.1,1$ \\ \cline{2-3}
 & \begin{tabular}[c]{@{}c@{}}KLDA \end{tabular} & $\delta=3,3.\overline{5},4.\overline{1},4.\overline{6},5.\overline{2},5.\overline{7},6.\overline{3},6.\overline{8},7.\overline{4},8$ \\ \hline
\textbf{Classifier} &Linear SVM & $c=10^{-8},\hdots,10^{-1},1,10$\\
\hline
\end{tabular}
\end{table}

%% file: Accuracy_Table_FG.tex
\begin{table}[h]
\centering
\caption{Gender Classification Results on FG-NET}
\label{table:classify_FG}
\begin{tabular}{|c|l|l|l|l|}
\hline
Method &Parameters &Acc\textsuperscript{(1)}
&TPR\textsuperscript{(2)} &TNR\textsuperscript{(3)} \\
\hline
\multirow{3}{*}{\begin{tabular}[c]{@{}c@{}}
\textbf{KPCA}\\ \end{tabular}}
 &$\delta=3.2\overline{6},c=10$ &\textbf{0.7025} &0.7325 &0.6621\\
&$\delta=3.\overline{3},c=10$ &0.6932 &0.7233 &0.6528\\
&$\delta=3.4,c=10$ &0.6801 &0.6651 &0.7001 \\
\hline
\multirow{3}{*}{\begin{tabular}[c]{@{}c@{}}\textbf{SKPCA} \end{tabular}}
 &$\delta=0.0098,\eta=0.1,c=10$ &\textbf{0.7154} &0.7542 &0.6633\\
&$\delta=0.0098,\eta=1,c=0.1$ &0.6933 &0.7413 &0.6289\\
&$\delta=0.0098,\eta=0.01,c=0.1$ &0.6893 &0.7701 &0.5809\\
\hline
\multirow{3}{*}{\begin{tabular}[c]{@{}c@{}}\textbf{KLDA} \end{tabular}}
&$\delta=3,c=0.01$ &\textbf{0.7225} &0.7593 &0.6730\\
&$\delta=5.\overline{7},c=1$ &0.7176 &0.7810 &0.6324\\
&$\delta=8,c=0.1$ &0.7131 &0.7431 &0.6727\\
\hline
\end{tabular}
\hspace{6cm}

\begin{minipage}{15cm}%
 \footnotesize 
(1) Acc represents mean accuracy. \\
(2) TPR represents mean true positive rate (recall/sensitivity): the proportion of male faces correctly classified. \\
(3) TNR represents mean true negative rate (specificity): the proportion of female faces correctly classified. \\
  \end{minipage}%
\end{table}
\normalsize

%% file: main.bbl
\begin{thebibliography}{-------}
\providecommand{\natexlab}[1]{#1}

\bibitem[Fodor(2002)]{fodor2002survey}
Fodor, I.K.
\newblock A survey of dimension reduction techniques.
\newblock Technical report, Lawrence Livermore National Lab., CA (US),  2002.

\bibitem[Sorzano \em{et~al.}(2014)Sorzano, Vargas, and
  Montano]{sorzano2014survey}
Sorzano, C.O.S.; Vargas, J.; Montano, A.P.
\newblock A survey of dimensionality reduction techniques.
\newblock {\em arXiv preprint arXiv:1403.2877} {\bf 2014}.

\bibitem[Izenman(2008)]{izenman2008modern}
Izenman, A.J.
\newblock Modern multivariate statistical techniques.
\newblock {\em Regression, classification and manifold learning} {\bf 2008}.

\bibitem[Pearson(1901)]{pearson1901liii}
Pearson, K.
\newblock LIII. On lines and planes of closest fit to systems of points in
  space.
\newblock {\em The London, Edinburgh, and Dublin Philosophical Magazine and
  Journal of Science} {\bf 1901}, {\em 2},~559--572.

\bibitem[Hotelling(1933)]{hotelling1933analysis}
Hotelling, H.
\newblock Analysis of a complex of statistical variables into principal
  components.
\newblock {\em Journal of educational psychology} {\bf 1933}, {\em 24},~417.

\bibitem[Yang \em{et~al.}(2003)Yang, Peng, Ward, and
  Rundensteiner]{yang2003interactive}
Yang, J.; Peng, W.; Ward, M.O.; Rundensteiner, E.A.
\newblock Interactive hierarchical dimension ordering, spacing and filtering
  for exploration of high dimensional datasets.
\newblock  IEEE Symposium on Information Visualization 2003 (IEEE Cat. No.
  03TH8714). IEEE,  2003, pp. 105--112.

\bibitem[Johansson and Johansson(2009)]{johansson2009interactive}
Johansson, S.; Johansson, J.
\newblock Interactive dimensionality reduction through user-defined
  combinations of quality metrics.
\newblock {\em IEEE transactions on visualization and computer graphics} {\bf
  2009}, {\em 15},~993--1000.

\bibitem[Fisher(1936)]{fisher1936use}
Fisher, R.A.
\newblock The use of multiple measurements in taxonomic problems.
\newblock {\em Annals of human genetics} {\bf 1936}, {\em 7},~179--188.

\bibitem[Rao(1948)]{rao1948utilization}
Rao, C.R.
\newblock The utilization of multiple measurements in problems of biological
  classification.
\newblock {\em Journal of the Royal Statistical Society. Series B
  (Methodological)} {\bf 1948}, {\em 10},~159--203.

\bibitem[Lee and Verleysen(2007)]{lee2007nonlinear}
Lee, J.A.; Verleysen, M.
\newblock {\em Nonlinear dimensionality reduction}; Springer Science \&
  Business Media,  2007.

\bibitem[Nhan~Duong \em{et~al.}(2015)Nhan~Duong, Luu, Gia~Quach, and
  Bui]{nhan2015beyond}
Nhan~Duong, C.; Luu, K.; Gia~Quach, K.; Bui, T.D.
\newblock Beyond principal components: Deep boltzmann machines for face
  modeling.
\newblock  Proceedings of the IEEE Conference on Computer Vision and Pattern
  Recognition,  2015, pp. 4786--4794.

\bibitem[Yin \em{et~al.}(2012)Yin, Liu, Jin, and Yang]{yin2012kernel}
Yin, J.; Liu, Z.; Jin, Z.; Yang, W.
\newblock Kernel sparse representation based classification.
\newblock {\em Neurocomputing} {\bf 2012}, {\em 77},~120--128.

\bibitem[Shawe-Taylor and Cristianini(2004)]{shawe2004kernel}
Shawe-Taylor, J.; Cristianini, N.
\newblock {\em Kernel methods for pattern analysis}; Cambridge university
  press,  2004.

\bibitem[Motai(2015)]{motai2015kernel}
Motai, Y.
\newblock Kernel association for classification and prediction: A survey.
\newblock {\em IEEE transactions on neural networks and learning systems} {\bf
  2015}, {\em 26},~208--223.

\bibitem[Xie \em{et~al.}(2012)Xie, Luu, and Savvides]{xie2012robust}
Xie, Y.; Luu, K.; Savvides, M.
\newblock A robust approach to facial ethnicity classification on large scale
  face databases.
\newblock  Biometrics: Theory, Applications and Systems (BTAS), 2012 IEEE Fifth
  International Conference on. IEEE,  2012, pp. 143--149.

\bibitem[Sch{\"o}lkopf \em{et~al.}(1997)Sch{\"o}lkopf, Smola, and
  M{\"u}ller]{scholkopf1997kernel}
Sch{\"o}lkopf, B.; Smola, A.; M{\"u}ller, K.R.
\newblock Kernel principal component analysis.
\newblock  International Conference on Artificial Neural Networks. Springer,
  1997, pp. 583--588.

\bibitem[Mika \em{et~al.}(1999)Mika, Ratsch, Weston, Scholkopf, and
  Mullers]{mika1999fisher}
Mika, S.; Ratsch, G.; Weston, J.; Scholkopf, B.; Mullers, K.R.
\newblock Fisher discriminant analysis with kernels.
\newblock  Neural networks for signal processing IX, 1999. Proceedings of the
  1999 IEEE signal processing society workshop. Ieee,  1999, pp. 41--48.

\bibitem[Baudat and Anouar(2000)]{baudat2000generalized}
Baudat, G.; Anouar, F.
\newblock Generalized discriminant analysis using a kernel approach.
\newblock {\em Neural computation} {\bf 2000}, {\em 12},~2385--2404.

\bibitem[Barzilay and Brailovsky(1999)]{barzilay1999domain}
Barzilay, O.; Brailovsky, V.L.
\newblock On domain knowledge and feature selection using a support vector
  machine.
\newblock {\em Pattern Recognition Letters} {\bf 1999}, {\em 20},~475--484.

\bibitem[Sch{\"o}lkopf \em{et~al.}(1998)Sch{\"o}lkopf, Simard, Smola, and
  Vapnik]{scholkopf1998prior}
Sch{\"o}lkopf, B.; Simard, P.; Smola, A.J.; Vapnik, V.
\newblock Prior knowledge in support vector kernels.
\newblock  Advances in neural information processing systems,  1998, pp.
  640--646.

\bibitem[Hinton and Salakhutdinov(2006)]{hinton2006reducing}
Hinton, G.E.; Salakhutdinov, R.R.
\newblock Reducing the dimensionality of data with neural networks.
\newblock {\em science} {\bf 2006}, {\em 313},~504--507.

\bibitem[Zhao \em{et~al.}(1998)Zhao, Krishnaswamy, Chellappa, Swets, and
  Weng]{zhao1998discriminant}
Zhao, W.; Krishnaswamy, A.; Chellappa, R.; Swets, D.L.; Weng, J.
\newblock Discriminant analysis of principal components for face recognition.
  In {\em Face Recognition}; Springer,  1998; pp. 73--85.

\bibitem[Mart{\'\i}nez and Kak(2001)]{martinez2001pca}
Mart{\'\i}nez, A.M.; Kak, A.C.
\newblock Pca versus lda.
\newblock {\em IEEE transactions on pattern analysis and machine intelligence}
  {\bf 2001}, {\em 23},~228--233.

\bibitem[Yang and Yang(2003)]{yang2003can}
Yang, J.; Yang, J.y.
\newblock Why can LDA be performed in PCA transformed space?
\newblock {\em Pattern recognition} {\bf 2003}, {\em 36},~563--566.

\bibitem[Turk and Pentland(1991)]{turk1991eigenfaces}
Turk, M.; Pentland, A.
\newblock Eigenfaces for recognition.
\newblock {\em Journal of cognitive neuroscience} {\bf 1991}, {\em 3},~71--86.

\bibitem[Belhumeur \em{et~al.}(1997)Belhumeur, Hespanha, and
  Kriegman]{belhumeur1997eigenfaces}
Belhumeur, P.N.; Hespanha, J.P.; Kriegman, D.J.
\newblock Eigenfaces vs. fisherfaces: Recognition using class specific linear
  projection.
\newblock {\em IEEE Transactions on pattern analysis and machine intelligence}
  {\bf 1997}, {\em 19},~711--720.

\bibitem[Kim \em{et~al.}(2002)Kim, Jung, and Kim]{kim2002face}
Kim, K.I.; Jung, K.; Kim, H.J.
\newblock Face recognition using kernel principal component analysis.
\newblock {\em IEEE signal processing letters} {\bf 2002}, {\em 9},~40--42.

\bibitem[Lu \em{et~al.}(2003)Lu, Plataniotis, and Venetsanopoulos]{lu2003face}
Lu, J.; Plataniotis, K.N.; Venetsanopoulos, A.N.
\newblock Face recognition using kernel direct discriminant analysis
  algorithms.
\newblock {\em IEEE Transactions on Neural Networks} {\bf 2003}, {\em
  14},~117--126.

\bibitem[Yang \em{et~al.}(2004)Yang, Zhang, Frangi, and Yang]{yang2004two}
Yang, J.; Zhang, D.; Frangi, A.F.; Yang, J.y.
\newblock Two-dimensional PCA: a new approach to appearance-based face
  representation and recognition.
\newblock {\em IEEE transactions on pattern analysis and machine intelligence}
  {\bf 2004}, {\em 26},~131--137.

\bibitem[Li and Yuan(2005)]{li20052d}
Li, M.; Yuan, B.
\newblock 2D-LDA: A statistical linear discriminant analysis for image matrix.
\newblock {\em Pattern Recognition Letters} {\bf 2005}, {\em 26},~527--532.

\bibitem[Karg \em{et~al.}(2009)Karg, Jenke, Seiberl, K{\"u}hnlenz, Schwirtz,
  and Buss]{karg2009comparison}
Karg, M.; Jenke, R.; Seiberl, W.; K{\"u}hnlenz, K.; Schwirtz, A.; Buss, M.
\newblock A comparison of PCA, KPCA and LDA for feature extraction to recognize
  affect in gait kinematics.
\newblock  Affective computing and intelligent interaction and workshops, 2009.
  ACII 2009. 3rd international conference on. IEEE,  2009, pp. 1--6.

\bibitem[Ye \em{et~al.}(2009)Ye, Shi, and Shi]{ye2009comparative}
Ye, F.; Shi, Z.; Shi, Z.
\newblock A comparative study of PCA, LDA and Kernel LDA for image
  classification.
\newblock  Ubiquitous Virtual Reality, 2009. ISUVR'09. International Symposium
  on. IEEE,  2009, pp. 51--54.

\bibitem[Yang \em{et~al.}(2004)Yang, Jin, Yang, Zhang, and
  Frangi]{yang2004essence}
Yang, J.; Jin, Z.; Yang, J.y.; Zhang, D.; Frangi, A.F.
\newblock Essence of kernel Fisher discriminant: KPCA plus LDA.
\newblock {\em Pattern Recognition} {\bf 2004}, {\em 37},~2097--2100.

\bibitem[Barshan \em{et~al.}(2011)Barshan, Ghodsi, Azimifar, and
  Jahromi]{barshan2011supervised}
Barshan, E.; Ghodsi, A.; Azimifar, Z.; Jahromi, M.Z.
\newblock Supervised principal component analysis: Visualization,
  classification and regression on subspaces and submanifolds.
\newblock {\em Pattern Recognition} {\bf 2011}, {\em 44},~1357--1371.

\bibitem[Wang \em{et~al.}(2015)Wang, Chen, Watkins, and
  Ricanek]{wang2015modified}
Wang, Y.; Chen, C.; Watkins, V.; Ricanek, K.
\newblock Modified Supervised Kernel PCA for Gender Classification.
\newblock  International Conference on Intelligent Science and Big Data
  Engineering. Springer,  2015, pp. 60--71.

\bibitem[Fewzee and Karray(2012)]{fewzee2012dimensionality}
Fewzee, P.; Karray, F.
\newblock Dimensionality reduction for emotional speech recognition.
\newblock  Privacy, Security, Risk and Trust (PASSAT), 2012 International
  Conference on and 2012 International Confernece on Social Computing
  (SocialCom). IEEE,  2012, pp. 532--537.

\bibitem[Samadani \em{et~al.}(2013)Samadani, Ghodsi, and
  Kuli{\'c}]{samadani2013discriminative}
Samadani, A.A.; Ghodsi, A.; Kuli{\'c}, D.
\newblock Discriminative functional analysis of human movements.
\newblock {\em Pattern Recognition Letters} {\bf 2013}, {\em 34},~1829--1839.

\bibitem[Wu \em{et~al.}(2013)Wu, Bowers, Huynh, and Souvenir]{wu2013biomedical}
Wu, H.; Bowers, D.M.; Huynh, T.T.; Souvenir, R.
\newblock Biomedical video denoising using supervised manifold learning.
\newblock  Biomedical Imaging (ISBI), 2013 IEEE 10th International Symposium
  on. IEEE,  2013, pp. 1244--1247.

\bibitem[Ashtiani and Ghodsi(2015)]{ashtiani2015dimension}
Ashtiani, H.; Ghodsi, A.
\newblock A dimension-independent generalization bound for kernel supervised
  principal component analysis.
\newblock  Feature Extraction: Modern Questions and Challenges,  2015, pp.
  19--29.

\bibitem[Fu \em{et~al.}(2010)Fu, Guo, and Huang]{fu2010age}
Fu, Y.; Guo, G.; Huang, T.S.
\newblock Age synthesis and estimation via faces: A survey.
\newblock {\em IEEE transactions on pattern analysis and machine intelligence}
  {\bf 2010}, {\em 32},~1955--1976.

\bibitem[Sun \em{et~al.}(2018)Sun, Zhang, Sun, and Tan]{sun2018demographic}
Sun, Y.; Zhang, M.; Sun, Z.; Tan, T.
\newblock Demographic analysis from biometric data: Achievements, challenges,
  and new frontiers.
\newblock {\em IEEE transactions on pattern analysis and machine intelligence}
  {\bf 2018}, {\em 40},~332--351.

\bibitem[Burton \em{et~al.}(1993)Burton, Bruce, and Dench]{burton1993s}
Burton, A.M.; Bruce, V.; Dench, N.
\newblock What's the difference between men and women? Evidence from facial
  measurement.
\newblock {\em Perception} {\bf 1993}, {\em 22},~153--176.

\bibitem[Ng \em{et~al.}(2012)Ng, Tay, and Goi]{ng2012vision}
Ng, C.B.; Tay, Y.H.; Goi, B.M.
\newblock Vision-based human gender recognition: A survey.
\newblock {\em arXiv preprint arXiv:1204.1611} {\bf 2012}.

\bibitem[Golomb \em{et~al.}(1990)Golomb, Lawrence, and
  Sejnowski]{golomb1990sexnet}
Golomb, B.A.; Lawrence, D.T.; Sejnowski, T.J.
\newblock Sexnet: A neural network identifies sex from human faces.
\newblock  NIPS,  1990, Vol.~1, p.~2.

\bibitem[Cottrell and Metcalfe(1991)]{cottrell1991empath}
Cottrell, G.W.; Metcalfe, J.
\newblock EMPATH: Face, emotion, and gender recognition using holons.
\newblock  Advances in neural information processing systems,  1991, pp.
  564--571.

\bibitem[Poggio \em{et~al.}(1992)Poggio, Brunelli, and
  Poggio]{poggio1992hyberbf}
Poggio, B.; Brunelli, R.; Poggio, T.
\newblock HyberBF networks for gender classification,  1992.

\bibitem[Wiskott \em{et~al.}(1995)Wiskott, Fellous, Kr{\"u}ger, and Von~der
  Malsburg]{wiskott1995face}
Wiskott, L.; Fellous, J.M.; Kr{\"u}ger, N.; Von~der Malsburg, C.
\newblock Face recognition and gender determination,  1995.

\bibitem[Guo and Mu(2010)]{guo2010human}
Guo, G.; Mu, G.
\newblock Human age estimation: What is the influence across race and gender?
\newblock  Computer Vision and Pattern Recognition Workshops (CVPRW), 2010 IEEE
  Computer Society Conference on. IEEE,  2010, pp. 71--78.

\bibitem[Guo and Mu(2011)]{guo2011simultaneous}
Guo, G.; Mu, G.
\newblock Simultaneous dimensionality reduction and human age estimation via
  kernel partial least squares regression.
\newblock  Computer vision and pattern recognition (cvpr), 2011 ieee conference
  on. IEEE,  2011, pp. 657--664.

\bibitem[Shan(2012)]{shan2012learning}
Shan, C.
\newblock Learning local binary patterns for gender classification on
  real-world face images.
\newblock {\em Pattern recognition letters} {\bf 2012}, {\em 33},~431--437.

\bibitem[Yang \em{et~al.}(2013)Yang, Lin, Chang, and Chen]{yang2013automatic}
Yang, H.F.; Lin, B.Y.; Chang, K.Y.; Chen, C.S.
\newblock Automatic age estimation from face images via deep ranking.
\newblock {\em networks} {\bf 2013}, {\em 35},~1872--1886.

\bibitem[Yi \em{et~al.}(2014)Yi, Lei, and Li]{yi2014age}
Yi, D.; Lei, Z.; Li, S.Z.
\newblock Age estimation by multi-scale convolutional network.
\newblock  Asian conference on computer vision. Springer,  2014, pp. 144--158.

\bibitem[Antipov \em{et~al.}(2016)Antipov, Berrani, and
  Dugelay]{antipov2016minimalistic}
Antipov, G.; Berrani, S.A.; Dugelay, J.L.
\newblock Minimalistic CNN-based ensemble model for gender prediction from face
  images.
\newblock {\em Pattern recognition letters} {\bf 2016}, {\em 70},~59--65.

\bibitem[Antipov \em{et~al.}(2017)Antipov, Baccouche, Berrani, and
  Dugelay]{antipov2017effective}
Antipov, G.; Baccouche, M.; Berrani, S.A.; Dugelay, J.L.
\newblock Effective training of convolutional neural networks for face-based
  gender and age prediction.
\newblock {\em Pattern Recognition} {\bf 2017}, {\em 72},~15--26.

\bibitem[Yang and Ai(2007)]{yang2007demographic}
Yang, Z.; Ai, H.
\newblock Demographic classification with local binary patterns.
\newblock  International Conference on Biometrics. Springer,  2007, pp.
  464--473.

\bibitem[Lian and Lu(2006)]{lian2006multi}
Lian, H.C.; Lu, B.L.
\newblock Multi-view gender classification using local binary patterns and
  support vector machines.
\newblock  International Symposium on Neural Networks. Springer,  2006, pp.
  202--209.

\bibitem[M{\"a}kinen and Raisamo(2008)]{makinen2008experimental}
M{\"a}kinen, E.; Raisamo, R.
\newblock An experimental comparison of gender classification methods.
\newblock {\em pattern recognition letters} {\bf 2008}, {\em 29},~1544--1556.

\bibitem[Alexandre(2010)]{alexandre2010gender}
Alexandre, L.A.
\newblock Gender recognition: A multiscale decision fusion approach.
\newblock {\em Pattern recognition letters} {\bf 2010}, {\em 31},~1422--1427.

\bibitem[Xia \em{et~al.}(2008)Xia, Sun, and Lu]{xia2008multi}
Xia, B.; Sun, H.; Lu, B.L.
\newblock Multi-view gender classification based on local Gabor binary mapping
  pattern and support vector machines.
\newblock  Neural Networks, 2008. IJCNN 2008.(IEEE World Congress on
  Computational Intelligence). IEEE International Joint Conference on. IEEE,
  2008, pp. 3388--3395.

\bibitem[Guo \em{et~al.}(2009)Guo, Dyer, Fu, and Huang]{guo2009gender}
Guo, G.; Dyer, C.R.; Fu, Y.; Huang, T.S.
\newblock Is gender recognition affected by age?
\newblock  Computer Vision Workshops (ICCV Workshops), 2009 IEEE 12th
  International Conference on. IEEE,  2009, pp. 2032--2039.

\bibitem[Han \em{et~al.}(2015)Han, Otto, Liu, and Jain]{han2015demographic}
Han, H.; Otto, C.; Liu, X.; Jain, A.K.
\newblock Demographic estimation from face images: Human vs. machine
  performance.
\newblock {\em IEEE Transactions on Pattern Analysis \& Machine Intelligence}
  {\bf 2015}, pp. 1148--1161.

\bibitem[Ma and Fu(2011)]{ma2011manifold}
Ma, Y.; Fu, Y.
\newblock {\em Manifold learning theory and applications}; CRC press,  2011.

\bibitem[Sch{\"o}lkopf \em{et~al.}(2001)Sch{\"o}lkopf, Herbrich, and
  Smola]{scholkopf2001generalized}
Sch{\"o}lkopf, B.; Herbrich, R.; Smola, A.J.
\newblock A generalized representer theorem.
\newblock  International conference on computational learning theory. Springer,
   2001, pp. 416--426.

\bibitem[Wahba(1990)]{wahba1990spline}
Wahba, G.
\newblock {\em Spline models for observational data}; Vol.~59, Siam,  1990.

\bibitem[Sch{\"o}lkopf \em{et~al.}(1998)Sch{\"o}lkopf, Smola, and
  M{\"u}ller]{scholkopf1998nonlinear}
Sch{\"o}lkopf, B.; Smola, A.; M{\"u}ller, K.R.
\newblock Nonlinear component analysis as a kernel eigenvalue problem.
\newblock {\em Neural computation} {\bf 1998}, {\em 10},~1299--1319.

\bibitem[Gretton \em{et~al.}(2005)Gretton, Bousquet, Smola, and
  Sch{\"o}lkopf]{gretton2005measuring}
Gretton, A.; Bousquet, O.; Smola, A.; Sch{\"o}lkopf, B.
\newblock Measuring statistical dependence with Hilbert-Schmidt norms.
\newblock  International conference on algorithmic learning theory. Springer,
  2005, pp. 63--77.

\bibitem[Steinwart and Christmann(2008)]{steinwart2008support}
Steinwart, I.; Christmann, A.
\newblock {\em Support vector machines}; Springer Science \& Business Media,
  2008.

\bibitem[Ricanek and Tesafaye(2006)]{ricanek2006morph}
Ricanek, K.; Tesafaye, T.
\newblock Morph: A longitudinal image database of normal adult age-progression.
\newblock  Automatic Face and Gesture Recognition, 2006. FGR 2006. 7th
  International Conference on. IEEE,  2006, pp. 341--345.

\bibitem[Japkowicz and Stephen(2002)]{japkowicz2002class}
Japkowicz, N.; Stephen, S.
\newblock The class imbalance problem: A systematic study.
\newblock {\em Intelligent data analysis} {\bf 2002}, {\em 6},~429--449.

\bibitem[Guo and Mu(2010)]{guo2010study}
Guo, G.; Mu, G.
\newblock A study of large-scale ethnicity estimation with gender and age
  variations.
\newblock  Computer Vision and Pattern Recognition Workshops (CVPRW), 2010 IEEE
  Computer Society Conference on. IEEE,  2010, pp. 79--86.

\bibitem[Yip \em{et~al.}(2018{\natexlab{a}})Yip, Bingham, Kempfert, Fabish,
  Kling, Chen, and Wang]{Yip2018preliminary}
Yip, B.; Bingham, G.; Kempfert, K.; Fabish, J.; Kling, T.; Chen, C.; Wang, Y.
\newblock Preliminary Studies on a Large Face Database.
\newblock {\em arXiv preprint arXiv:1811.06446} {\bf 2018}.

\bibitem[Yip \em{et~al.}(2018{\natexlab{b}})Yip, Towner, Kling, Chen, and
  Wang]{Yip2018preprocess}
Yip, B.; Towner, R.; Kling, T.; Chen, C.; Wang, Y.
\newblock Image Pre-processing Using OpenCV Library on MORPH-II Face Database.
\newblock {\em arXiv preprint arXiv:1811.06934} {\bf 2018}.

\bibitem[Edwards \em{et~al.}(1998)Edwards, Taylor, and
  Cootes]{edwards1998interpreting}
Edwards, G.J.; Taylor, C.J.; Cootes, T.F.
\newblock Interpreting face images using active appearance models.
\newblock  Proceedings Third IEEE International Conference on Automatic Face
  and Gesture Recognition. IEEE,  1998, pp. 300--305.

\bibitem[Kling(2017)]{feature}
Kling, T.
\newblock Morph-II: Feature Vector Documentation: NSF-REU site at UNC
  Wilmington.
\newblock \url{http://libres.uncg.edu/ir/uncw/f/wangy2018-1.pdf},  2017.

\bibitem[Byun and Lee(2002)]{byun2002applications}
Byun, H.; Lee, S.W.
\newblock Applications of support vector machines for pattern recognition: A
  survey. In {\em Pattern recognition with support vector machines}; Springer,
  2002; pp. 213--236.

\bibitem[Qiu \em{et~al.}(2016)Qiu, Mei, and Qiu]{qiu2016package}
Qiu, Y.; Mei, J.; Qiu, M.Y.
\newblock Package ‘rARPACK’ {\bf 2016}.

\bibitem[Dimitriadou \em{et~al.}(2005)Dimitriadou, Hornik, Leisch, Meyer, and
  Weingessel]{dimitriadou2005misc}
Dimitriadou, E.; Hornik, K.; Leisch, F.; Meyer, D.; Weingessel, A.
\newblock Misc Functions of the Department of Statistics (e1071), TU Wien.
\newblock {\em R package version} {\bf 2005}, pp. 1--5.

\bibitem[Platt \em{et~al.}(1999)Platt et~al.]{platt1999probabilistic}
Platt, J.; others.
\newblock Probabilistic outputs for support vector machines and comparisons to
  regularized likelihood methods.
\newblock {\em Advances in large margin classifiers} {\bf 1999}, {\em
  10},~61--74.

\bibitem[Guo and Mu(2014)]{guo2014framework}
Guo, G.; Mu, G.
\newblock A framework for joint estimation of age, gender and ethnicity on a
  large database.
\newblock {\em Image and Vision Computing} {\bf 2014}, {\em 32},~761--770.

\bibitem[Panis \em{et~al.}(2016)Panis, Lanitis, Tsapatsoulis, and
  Cootes]{panis2016overview}
Panis, G.; Lanitis, A.; Tsapatsoulis, N.; Cootes, T.F.
\newblock Overview of research on facial ageing using the FG-NET ageing
  database.
\newblock {\em IET Biometrics} {\bf 2016}, {\em 5},~37--46.

\bibitem[Cootes \em{et~al.}(1998)Cootes, Edwards, and Taylor]{cootes1998active}
Cootes, T.F.; Edwards, G.J.; Taylor, C.J.
\newblock Active appearance models.
\newblock  European conference on computer vision. Springer,  1998, pp.
  484--498.

\bibitem[Wang \em{et~al.}(2010)Wang, Ricanek, Chen, and Chang]{wang2010gender}
Wang, Y.; Ricanek, K.; Chen, C.; Chang, Y.
\newblock Gender classification from infants to seniors.
\newblock  2010 Fourth IEEE International Conference on Biometrics: Theory,
  Applications and Systems (BTAS). IEEE,  2010, pp. 1--6.

\end{thebibliography}
